\PassOptionsToPackage{table}{xcolor}
\documentclass[letterpaper]{article} 
\usepackage{aaai2027}  
\usepackage[hyphens]{url}  
\usepackage{graphicx} 
\usepackage{natbib}  
\usepackage{caption} 
\usepackage{booktabs}

\usepackage{latexsym}
\usepackage[utf8]{inputenc}
\usepackage{microtype}
\usepackage[most]{tcolorbox} 
\usepackage{amssymb}
\usepackage{csquotes}
\usepackage{multirow}
\usepackage{arydshln}

\usepackage{pifont} 

\definecolor{seekerbg}{HTML}{F3F4F6}
\definecolor{seekerborder}{HTML}{E5E7EB}
\definecolor{basebg}{HTML}{FEF2F2}
\definecolor{baseborder}{HTML}{FECACA}
\definecolor{basefont}{HTML}{7F1D1D}
\definecolor{ourbg}{HTML}{F0FDF4}
\definecolor{ourborder}{HTML}{BBF7D0}
\definecolor{ourfont}{HTML}{14532D}
\definecolor{thinkbg}{HTML}{EFF6FF}
\definecolor{thinkborder}{HTML}{93C5FD}
\definecolor{thinkfont}{HTML}{1E3A8A}

\newcommand{\xmark}{\ding{55}}
\newcommand{\cmark}{\ding{51}}

\usepackage{listings}
\usepackage{tabularx}
\usepackage{subcaption}

\definecolor{caseblue}{HTML}{1D4ED8}
\definecolor{casebluebg}{HTML}{EFF6FF}
\definecolor{casegreen}{HTML}{166534}
\definecolor{casegreenbg}{HTML}{F0FDF4}
\definecolor{casemuted}{HTML}{64748B}
\definecolor{casefailure}{HTML}{991B1B}
\definecolor{casefailurebg}{HTML}{FEF2F2}

\lstdefinestyle{prompt}{
    basicstyle=\tiny\fontfamily{pcr}\selectfont,
    breaklines=true,
    breakatwhitespace=false,
    columns=fullflexible,
    keepspaces=true,
    showstringspaces=false,
    inputencoding=utf8,
    literate={—}{{--}}1 {é}{{\'e}}1
}

\usepackage{cleveref}  

\title{CounselPlex: A Stressor-Aware Full-Duplex Counseling Model}

\author {
    Yeonju Kim\textsuperscript{\rm 1},
    Bella Godiva\textsuperscript{\rm 1},
    Yong Man Ro\textsuperscript{\rm 1}\thanks{Corresponding author.}
}
\affiliations {
    \textsuperscript{\rm 1}Integrated Vision and Language Lab, KAIST, South Korea \\
    \{yeonju7.kim, bellagodiva, ymro\}@kaist.ac.kr
}

\nocopyright
\begin{document}

\maketitle

\begin{abstract}
Full-duplex speech models can listen and speak at the same time, enabling natural, low-latency interaction. In counseling, however, these models often focus on the latest utterance and lose track of the seeker's underlying problem. We call this failure \textbf{Stressor Blindness}. We propose \textbf{CounselPlex}, a stressor-aware full-duplex framework that uses two operations. CounselPlex updates a short stressor state during seeker silence and injects this state into \texttt{<PAD>} positions while listening. We also introduce RoPE Position Shift Augmentation (RPSA), which broadens positional coverage during training and improves performance at longer turn depths in our evaluation. LLM-judge evaluations show improved counseling quality and stressor grounding over strong baselines. 
\end{abstract}

\section{Introduction}
\label{sec:intro}

\begin{figure}[t!]
\centering
\begin{tcolorbox}[
    enhanced,
    colback=white,
    colframe=gray!45,
    boxrule=0.6pt,
    arc=2pt,
    left=3pt,
    right=3pt,
    top=3pt,
    bottom=4pt,
    boxsep=1pt
]

\centering
\textbf{\small Stressor Blindness: Does the Model Track the Stressor?}\par

\begin{tcolorbox}[
    colback=seekerbg,
    colframe=seekerborder,
    boxrule=0.5pt,
    arc=1.5pt,
    left=3pt,
    right=3pt,
    top=2pt,
    bottom=2pt,
    width=0.96\linewidth,
    boxsep=0pt
]
\small
\textbf{Seeker ($t{-}8$), initial disclosure:}
I've experienced \textbf{trauma} recently, and I'm really
struggling to cope.
\end{tcolorbox}

\begin{tcolorbox}[
    colback=thinkbg,
    colframe=thinkborder,
    coltext=thinkfont,
    boxrule=0.5pt,
    arc=1.5pt,
    left=3pt,
    right=3pt,
    top=1.5pt,
    bottom=1.5pt,
    width=0.94\linewidth,
    boxsep=0pt
]
\small
\centering
\textbf{Silence update:}
\texttt{<think> trauma </think>}
\end{tcolorbox}

{\small\itshape\color{gray}
\(\downarrow\)\quad stored and re-injected while the model listens
\quad\(\downarrow\)\par}
{\small\itshape\color{gray}
\ldots{} eight turns later \ldots{}\par}

\begin{tcolorbox}[
    colback=seekerbg,
    colframe=seekerborder,
    boxrule=0.5pt,
    arc=1.5pt,
    left=3pt,
    right=3pt,
    top=2pt,
    bottom=2pt,
    width=0.96\linewidth,
    boxsep=0pt
]
\small
\textbf{Seeker (current):}
I feel like I'm constantly on edge, like something bad is
going to happen any moment.
\end{tcolorbox}

\begin{tcolorbox}[
    colback=basebg,
    colframe=baseborder,
    coltext=basefont,
    boxrule=0.5pt,
    arc=1.5pt,
    left=3pt,
    right=3pt,
    top=2pt,
    bottom=2pt,
    width=0.96\linewidth,
    boxsep=0pt
]
\small
\textbf{PersonaPlex\textsuperscript{\textdagger} (baseline):}
\ldots{} a lot of \textbf{tension and anxiety}. That can be
really tough to deal with. There are some
\textbf{mindfulness techniques} \ldots{}
\end{tcolorbox}

{\small\hspace{0.02\linewidth}\parbox{0.94\linewidth}{
\textcolor{red}{\xmark~\textbf{Generic support:}}
Acknowledges the anxiety but misses the earlier trauma.}}\par

\begin{tcolorbox}[
    colback=ourbg,
    colframe=ourborder,
    coltext=ourfont,
    boxrule=0.5pt,
    arc=1.5pt,
    left=3pt,
    right=3pt,
    top=2pt,
    bottom=2pt,
    width=0.96\linewidth,
    boxsep=0pt
]
\small
\textbf{CounselPlex (ours):}
It sounds like \textbf{your trauma} has been making you feel
like that. Would you like to talk more about
\textbf{your trauma}?
\end{tcolorbox}

{\small\hspace{0.02\linewidth}\parbox{0.94\linewidth}{
\textcolor{green!55!black}{\cmark~\textbf{Stressor connected:}}
Links the current anxiety back to the earlier trauma.}}\par

\end{tcolorbox}

\caption{\textbf{Held-out example of Stressor Blindness.}
Trauma was disclosed eight turns before the seeker said they felt
``constantly on edge.'' The fine-tuned
PersonaPlex\textsuperscript{\textdagger} baseline responds only to
this symptom. CounselPlex links the feeling back to
the earlier trauma. The baseline output is shortened only where
ellipses appear.}
\label{fig:stressor_blindness}
\end{figure}

Recent full-duplex speech models have expanded from generic
assistants~\cite{defossez2024moshi} to role-conditioned conversational
agents~\cite{roy2026personaplex}.
For example, PersonaPlex combines role and
voice conditioning with concurrent listening and speaking. A single
model can therefore take on specialized spoken roles.
This capability supports interaction-intensive applications such as
customer service and counseling-oriented emotional support. In these
applications, natural conversational flow can support
rapport~\cite{gratch2007creating}, while the agent must also behave
consistently with its role. However, role conditioning tells
an agent how to behave, not which user-specific information to retain
as the conversation develops.

This limitation is especially important in counseling. A seeker may
first disclose a stressor and mention only one of its symptoms many
turns later. Figure~\ref{fig:stressor_blindness} shows a held-out
example. Eight turns after the seeker disclosed trauma, a
role-conditioned full-duplex baseline treats feeling ``constantly on
edge'' as general anxiety. CounselPlex instead links this feeling back
to the earlier trauma. We call the baseline failure \textbf{Stressor
Blindness}: the inability to maintain and use the seeker's active
stressor state across turns. This failure often appears as generic
support: the response sounds caring but does not connect the current
symptom to the earlier stressor.



Clinical and neuroscientific research suggests that addressing an
underlying stressor, rather than only its immediate symptoms, supports
recovery and emotional resilience~\cite{beck2020cognitive,creswell2007neural,lazarus1984stress}.
A counseling model should therefore retain the stressor while
responding in real time.

We call this capability \textbf{Stressor-Aware Understanding}: retaining
the stressor throughout a conversation and using it to interpret
immediate symptoms. In our framework, the stressor state is a concise
1--7 word noun phrase that describes the seeker's underlying problem
(e.g., \enquote{bereavement}).

We propose \textbf{CounselPlex}, a full-duplex framework with two main
operations. First, it detects changes in the seeker's situation and
generates a new stressor state when needed. Second, it injects the
active state into the agent's text stream while the agent listens.
The model can then directly use the stressor to interpret each new
utterance. To improve stability in long conversations, we also use
RoPE Position Shift Augmentation (RPSA) during training.

We evaluate CounselPlex for both counseling quality and full-duplex
behavior. GPT-4.1~\cite{achiam2023gpt} scores responses from 1 to 5 on
empathy, coherence, context recalling, strategy appropriateness, and
stressor awareness. We also measure turn-taking, latency, and overlap
handling. Finally, we analyze when the model generates and uses
stressor states.

Our main contributions are as follows:
\begin{itemize}
    \item \textbf{CounselPlex:} a full-duplex framework that maintains therapeutic context by generating and injecting short stressor states.
    \item \textbf{RPSA:} a training method that broadens positional coverage and improves long-context stability.
    \item \textbf{Evaluation:} experiments with LLM judges, duplex benchmarks, and stressor-state analyses that show improved counseling quality with a modest latency increase over the fine-tuned baseline.
\end{itemize}

\section{Related Work}
\label{sec:related_works}
\subsection{Emotional Support Conversation}
Emotional Support Conversation (ESC) research has progressed from
rule-based support agents~\cite{van2012bdi,medeiros2018using} to
neural models for emotional and empathetic
dialogue~\cite{zhou2018emotional,zhou2018mojitalk,
rashkin2019empatheticopendomainconversationmodels,lin2019moel}.
ESConv then modeled support as a sequence of explicit strategies, such
as asking questions and reflecting feelings~\cite{liu2021towards}.
These strategies overlap with established counseling
skills~\cite{burleson2003emotional,hill1999helping}.
Recent LLM systems extend empathy and multi-turn
support~\cite{chen2023soulchat,liu2023chatcounselor,lai2023psy},
while data augmentation and explicit reasoning improve strategy
use~\cite{zheng2023augesc,qiu2024smile,zhang2024escot}.

Most of these systems are text-based and assume strict turn taking:
the supporter receives a complete utterance before responding.
They therefore do not model real-time speech events such as overlap,
interruption, or backchannels, although interaction timing can affect
rapport~\cite{gratch2007creating}. CounselPlex instead focuses on
tracking counseling context in a full-duplex spoken setting.

\subsection{Full-Duplex Speech Language Models}
Full-duplex speech language models (FD-SLMs) listen and speak
simultaneously. Moshi, Freeze-Omni, Mini-Omni, and Llama-Omni are
end-to-end examples~\cite{defossez2024moshi,wang2024freeze,
xie2024mini,fang2025llama}. Subsequent work improves codec-free
streaming, interaction control, and persona-consistent emotional
prosody~\cite{yu2024salmonn,zufle2026f,roy2026personaplex}.

Benchmarks measure pauses, overlap, turn taking, and
interruptions~\cite{lin2025full,lin2026full}. Multi-round evaluations
show that performance declines as conversations
lengthen~\cite{zhang2025mtr,
lin2026fullduplexbenchv2multiturnevaluationframework}. This creates a
need for methods that retain context without disrupting real-time
interaction. One option is a cascaded system such as FireRedChat,
which supports external modules but may add
latency~\cite{chen2025fireredchat}. CounselPlex instead remains
end-to-end: it injects a compact stressor state directly into the text
stream to retain counseling context.

\begin{figure*}[t]
\centering		
\centerline{\includegraphics[width=1.0\linewidth]{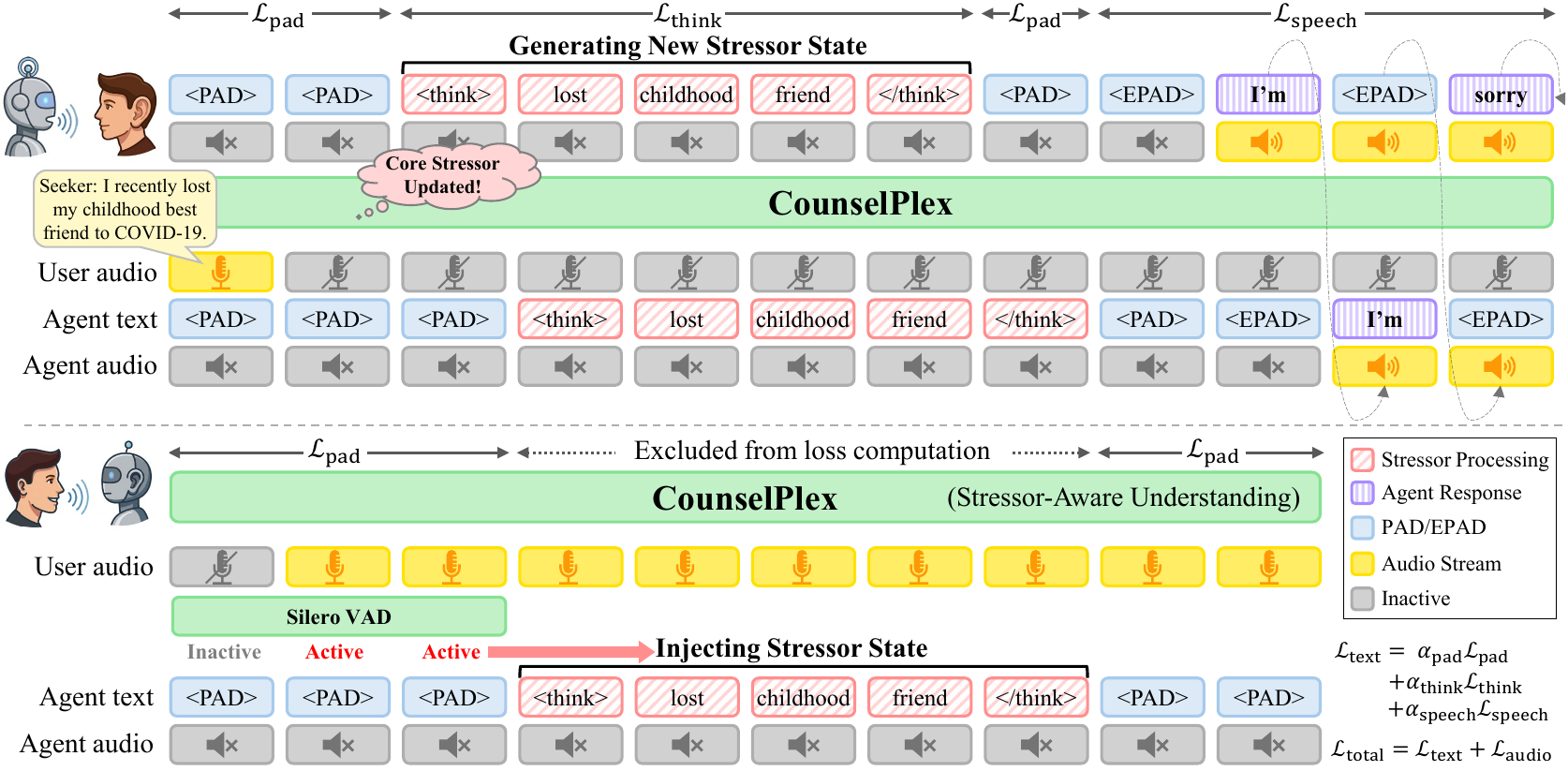}}
\caption{Overview of the CounselPlex framework. During seeker silence (top), CounselPlex generates an updated stressor state as a short think trace. When the seeker resumes speaking (bottom), the framework force-injects this state into \texttt{<PAD>} slots so that the agent can use it to interpret the incoming utterance.}
\label{fig:overall_architecture}
\end{figure*}

\section{CounselPlex: Stressor-State Augmented Full-Duplex Model}
\label{sec:method}

\subsection{Data Construction \& Stressor Labeling}
\label{subsec:data}
The original ESConv~\cite{liu2021towards} and ESCoT~\cite{zhang2024escot} datasets contain text only, so we convert them into spoken dialogues. We synthesize the seeker and supporter turns with CosyVoice3~\cite{du2025cosyvoice}, using one male voice for all seekers and one female voice for all supporters. This process produces 80,034 utterance-level WAV files sampled at 24 kHz. The corpus contains 3,009 conversations and 158.3 hours of speech across all splits.

For training, we annotate each conversation with a core stressor. GPT-4.1~\cite{achiam2023gpt} extracts a 1--7 word phrase that summarizes the seeker's underlying situation. The prompt asks the model to use factual disclosures rather than temporary symptoms (e.g., \enquote{unemployed, facing bankruptcy} or \enquote{raising children alone}). We omit this label when the seeker gives no external stressor, as in some conversations about depression. In total, we leave 127 conversations (4.22\%) without a stressor annotation. Appendix~\ref{appendix:stress_labeling_prompt} provides examples and the full prompt.
\subsection{Overall Architecture}
\label{sec:overall_arch}
Figure~\ref{fig:overall_architecture} shows the two main operations in CounselPlex. The model first generates or updates a stressor state, then injects the active state while listening to a later seeker turn. We use PersonaPlex~\cite{roy2026personaplex}, which builds on Moshi~\cite{defossez2024moshi}, as the backbone because its separate streams support simultaneous listening and speaking.

\subsubsection{Generating New Stressor State}
\label{subsubsec:generation_stressor}
CounselPlex learns to infer a core stressor from the dialogue history. Before generating a stressor, it waits for 0.5 seconds of seeker silence, treating this as utterance completion. This threshold prevents short pauses in the middle of a sentence from being mistaken for the end of a turn. The model then uses the dialogue history to decide whether to generate a new stressor state as a short think trace. If no update is needed, the current state remains active.

\subsubsection{Injecting Stressor State}
Generation and injection use different triggers. The generated stressor remains active until the model produces a new one. When the seeker starts a later turn, Silero VAD v5~\cite{SileroVAD} checks for sufficient speech activity within a 0.5-second window. Once speech is confirmed, CounselPlex force-injects the active stressor state into the agent's \texttt{<PAD>} positions. Requiring enough active frames prevents brief backchannels and acoustic noise from being treated as full turns.

\subsection{Model Training}

\subsubsection{Total Loss}
The overall training objective $\mathcal{L}_{\text{total}}$ is the sum of the text and audio losses:
\begin{equation}
    \mathcal{L}_{\text{total}} = \mathcal{L}_{\text{text}} + \mathcal{L}_{\text{audio}}
\end{equation}

\paragraph{Text Loss Decomposition.}
We divide the agent's generated text tokens into three groups. \enquote{Think} tokens form the stressor trace between \texttt{<think>} and \texttt{</think>}. \enquote{Pad} tokens are the \texttt{<PAD>} tokens that represent silence. \enquote{Speech} tokens contain the spoken words and their \texttt{<EPAD>} tokens, where an \texttt{<EPAD>} token appears immediately before each spoken word. We compute token-level cross-entropy for each group and combine the three losses with separate weights:
\begin{equation}
\mathcal{L}_{\text{text}} = \alpha_{\text{pad}} \mathcal{L}_{\text{pad}} + \alpha_{\text{think}} \mathcal{L}_{\text{think}} + \alpha_{\text{speech}} \mathcal{L}_{\text{speech}}
\end{equation}
Separate weights are necessary because silence produces far more \texttt{<PAD>} tokens than the other groups. Stressor traces are also sparse, with 1.19 stressors per conversation on average. Giving \textit{think} and \textit{speech} equal weights causes the model to generate stressor states too often, so we set $\alpha_{\text{think}}$ lower than $\alpha_{\text{speech}}$. We exclude the injection window from the text loss because the tokens in this window are inserted by the system rather than generated by the model. Although we apply no text loss in this window, the model generally predicts \texttt{<PAD>} and remains silent because \texttt{<PAD>} is the most common token in the training data.

\paragraph{Audio Loss Calculation.}
The model processes the user and agent audio streams at the same time, but $\mathcal{L}_{\text{audio}}$ supervises only the generated agent stream. Following Moshi~\cite{defossez2024moshi}, we represent each audio frame with 8 discrete tokens, one from each Mimi codec codebook. We average the cross-entropy loss over all codebooks and all $T$ generated audio frames:
\begin{equation}
    \mathcal{L}_{\text{audio}} = \frac{1}{8T} \sum_{t=1}^{T} \sum_{c=1}^{8} \text{CE}(\hat{a}_{t,c}, a_{t,c})
\end{equation}
Here, $a_{t,c}$ is the ground-truth acoustic token at frame $t$ and codebook $c$, and $\hat{a}_{t,c}$ is the predicted token distribution at the same position. $\text{CE}(\cdot, \cdot)$ denotes cross-entropy between the prediction and target.

\subsubsection{RoPE Position Shift Augmentation (RPSA) for Long-Context Stability}
Counseling conversations can exceed the model's fixed training window. Most training examples are shorter, so the model sees low Rotary Position Embedding (RoPE)~\cite{su2024roformer} indices much more often than high indices. This imbalance lowers generation quality in long evaluation dialogues.

RPSA exposes the model to a wider range of positions during training. For a conversation that exceeds the context window, we first crop a fixed-size segment. If the segment begins after 750 frames (approximately 60 seconds), we add a random offset to its starting RoPE index with probability 0.7 and shift every position in the segment by the same amount. Thus, RPSA changes the segment's absolute positions but preserves the order and relative spacing of the dialogue. Section~\ref{subsec:rpsa_effect} analyzes its effect on long conversations.

\subsection{Real-Time Inference}
\label{sec:inference}
At deployment, CounselPlex continuously processes the seeker's audio as a real-time stream. Stressor generation and injection follow separate triggers. After a seeker turn ends, the model uses the dialogue context to decide whether to generate a new stressor and, if so, stores it as the active state. Later, Silero VAD~\cite{SileroVAD} confirms the start of the next valid seeker turn by detecting sufficient speech activity in a 0.5-second window. The system then force-injects the stored state into the agent's text stream; it does not regenerate the state at this point. This separation lets the model update the state during silence and apply it while interpreting the next utterance. The Mimi decoder continuously converts the agent's generated acoustic tokens into a $24\,\text{kHz}$ audio waveform.

\section{Experimental Setup}
\label{sec:experiments}
\subsection{Implementation Details}
\label{subsec:impl_details}
\begin{table*}[!t]
\centering
\caption{Counseling quality on the ESConv and ESCoT test sets, evaluated by GPT-4.1. Avg5 is the utterance-level mean of Empathy, Coherence, Context Recalling, Strategy Appropriateness, and Stressor Awareness. RR is the Response Rate. PersonaPlex† is the backbone fine-tuned on our data without stressor reasoning. Bold marks the best model result in each column, excluding human GT. Higher is better.}
\resizebox{\linewidth}{!}{
\begin{tabular}{lcccccccccccccc}
\hline
\multirow{3}{*}{Model} & \multicolumn{7}{c}{\textbf{ESConv} (194 conv, 2,286 turns)} & \multicolumn{7}{c}{\textbf{ESCoT} (341 conv, 3,992 turns)} \\ \cmidrule(lr){2-8} \cmidrule(lr){9-15}
& RR & Empathy & Coherence & Context & Strategy & Stressor & \textbf{Avg5} 
& RR & Empathy & Coherence & Context & Strategy & Stressor & \textbf{Avg5} \\ \hline
Moshi & 0.809 & 2.155 & 2.288 & 2.259 & 2.160 & 2.294 & 2.231 & 0.785 & 2.203 & 2.323 & 2.309 & 2.213 & 2.348 & 2.279 \\ 
PersonaPlex & 0.840 & 2.198 & 2.455 & 2.384 & 2.213 & 2.398 & 2.329 & 0.671 & 2.113 & 2.384 & 2.288 & 2.160 & 2.300 & 2.249 \\ 
PersonaPlex$^\dagger$ & \textbf{0.960} & 2.038 & 2.210 & 2.206 & 2.056 & 2.204 & 2.143 & \textbf{0.939} & 2.498 & 2.740 & 2.736 & 2.525 & 2.720 & 2.644 \\ 
\cellcolor[HTML]{F0F0F0}CounselPlex & \cellcolor[HTML]{F0F0F0}0.955 & \cellcolor[HTML]{F0F0F0}\textbf{2.363} & \cellcolor[HTML]{F0F0F0}\textbf{2.600} & \cellcolor[HTML]{F0F0F0}\textbf{2.546} & \cellcolor[HTML]{F0F0F0}\textbf{2.376} & \cellcolor[HTML]{F0F0F0}\textbf{2.521} & \cellcolor[HTML]{F0F0F0}\textbf{2.481} & \cellcolor[HTML]{F0F0F0}0.929 & \cellcolor[HTML]{F0F0F0}\textbf{2.654} & \cellcolor[HTML]{F0F0F0}\textbf{2.910} & \cellcolor[HTML]{F0F0F0}\textbf{2.902} & \cellcolor[HTML]{F0F0F0}\textbf{2.693} & \cellcolor[HTML]{F0F0F0}\textbf{2.906} & \cellcolor[HTML]{F0F0F0}\textbf{2.813} \\ \hdashline
GT (human) & 1.000 & 2.626 & 3.117 & 2.989 & 2.657 & 2.893 & 2.856 & 1.000 & 3.161 & 3.714 & 3.594 & 3.303 & 3.585 & 3.471 \\ \hline
\end{tabular}
\label{table_main_results}
}
\end{table*}

\begin{table*}[!t]
\centering
\caption{Real-time behavior on Full-Duplex-Bench~\citep{lin2025full}. TOR is the fraction of examples in which the model takes a full conversational turn rather than remaining silent or producing only a backchannel; the preferred direction depends on the task. Freq is backchannel frequency, and JSD is the Jensen--Shannon divergence between predicted and reference backchannel distributions. GPT-4o scores interruption responses from 0 to 5. Arrows show the preferred direction; bold marks the best result in each column.}
\label{tab:full_duplex_bench}
\resizebox{\textwidth}{!}{
\begin{tabular}{lcccccccccc}
\toprule
& \multicolumn{1}{c}{\textbf{Pause (Synthetic)}} & \multicolumn{1}{c}{\textbf{Pause (Candor)}} & \multicolumn{3}{c}{\textbf{Backchannel}} & \multicolumn{2}{c}{\textbf{Smooth Turn Taking}} & \multicolumn{3}{c}{\textbf{User Interruption}} \\
\cmidrule(lr){2-3} \cmidrule(lr){4-6} \cmidrule(lr){7-8} \cmidrule(lr){9-11}
 \textbf{Model} & TOR (↓) & TOR (↓) & TOR (↓) & Freq (↑) & JSD (↓) & TOR (↑) & Latency (↓) & TOR (↑) & GPT-4o (↑) & Latency (↓) \\
\midrule
Moshi        & 0.985 & 0.980 & 1.000 & 0.001 & 0.957 & 0.941 & 0.265 & 1.000 & 0.765 & \textbf{0.257} \\
PersonaPlex  & 0.584 & 0.662 & 0.327 & \textbf{0.025} & \textbf{0.649} & 0.992 & \textbf{0.070} & \textbf{1.000} & \textbf{4.210} & 0.400 \\
PersonaPlex$^{\dagger}$ & 0.248 & 0.370 & 0.236 & 0.009 & 0.950 & 0.950 & 0.920 & 0.290 & 2.931 & 0.971 \\
\cellcolor[HTML]{F0F0F0}CounselPlex & \cellcolor[HTML]{F0F0F0}\textbf{0.066} & \cellcolor[HTML]{F0F0F0}\textbf{0.213} & \cellcolor[HTML]{F0F0F0}\textbf{0.000} & \cellcolor[HTML]{F0F0F0}0.000 & \cellcolor[HTML]{F0F0F0}1.000 & \cellcolor[HTML]{F0F0F0}\textbf{0.992} & \cellcolor[HTML]{F0F0F0}0.988 & \cellcolor[HTML]{F0F0F0}0.990 & \cellcolor[HTML]{F0F0F0}3.071 & \cellcolor[HTML]{F0F0F0}1.056 \\
\hline
\end{tabular}
}
\end{table*}

We use PersonaPlex~\cite{roy2026personaplex} as our baseline and fine-tune it on a synthesized spoken corpus. The corpus is derived from two text-based therapeutic datasets, \textbf{ESConv}~\cite{liu2021towards} and \textbf{ESCoT}~\cite{zhang2024escot}, both of which focus on strategy-driven emotional support conversations. It contains 158.3 hours of speech across 3,009 sessions: 2,107 training, 367 validation, and 535 test dialogues. A dialogue lasts 3.1 minutes on average and up to 13.0 minutes. The average turn duration is 16.5 seconds, including silence between speakers.

CounselPlex has approximately 8.4B parameters and inherits the PersonaPlex/Moshi backbone architecture. We fine-tune it for 4 epochs on four NVIDIA RTX A6000 (48GB) GPUs. Training uses 8-bit AdamW~\cite{loshchilov2017decoupled}, DeepSpeed ZeRO Stage 2~\cite{rajbhandari2020zero}, a batch size of 16, and a peak learning rate of $1\times10^{-5}$. Fine-tuning costs approximately 45 GPU-hours ($\approx$ 11.2 wall-clock hours), excluding the cost of pre-training the PersonaPlex backbone. We conduct a single training run. All results are from this single model.

Following Moshi~\cite{defossez2024moshi}, we use a 12.5 fps Mimi frame rate and a 2048-frame context. We set the text-loss weights to $\alpha_{\text{pad}}=0.3$, $\alpha_{\text{think}}=0.5$, and $\alpha_{\text{speech}}=1.5$. For RPSA, long-conversation crops that start after 750 frames ($\approx$ 60 s) receive a random offset ($\Delta \sim \text{Uniform}\{1000, 4000\}$) with probability 0.7. We also add a random pause of $0$--$5$ seconds after each supporter turn in the training data. These pauses teach the model to remain silent until the seeker speaks again.

\subsection{Baseline Models \& Evaluation Setup}
\label{subsec:baselines_evalbench}
We compare CounselPlex with three end-to-end full-duplex baselines. \textbf{Moshi}~\cite{defossez2024moshi} is a foundational full-duplex model with dual-stream audio and Inner Monologue. \textbf{PersonaPlex}~\cite{roy2026personaplex} is our zero-shot backbone without stressor tracking. \textbf{PersonaPlex}\textsuperscript{\textdagger} is the same backbone fine-tuned on our data, but without stressor reasoning. This last comparison separates the effect of stressor reasoning from that of in-domain fine-tuning. We exclude cascade and frozen-LLM pipelines because their latency is not comparable and they do not support bidirectional streaming. We evaluate both counseling quality and full-duplex behavior.

\paragraph{Counseling Quality Assessment.}
Standard $n$-gram metrics, such as BLEU, are inadequate for open-ended therapeutic dialogue. We therefore use GPT-4.1 as an LLM-as-a-judge~\cite{zheng2023judging} to score the responses. The full prompt and rubric appear in Appendix~\ref{appendix:llm_evaluation_prompt}. To avoid transcription noise introduced by ASR, we evaluate the model's text-stream output directly rather than transcribing its synthesized speech.

The judge receives the prior dialogue history, the seeker's current utterance, and the supporter's response. It rates the response on a strict 1--5 scale along five dimensions: \textit{Empathy}; \textit{Coherence}, or responsiveness to the latest turn; \textit{Context Recalling}, or consistency with the dialogue history; \textit{Strategy Appropriateness}, or selecting a context-appropriate strategy without giving premature suggestions; and \textit{Stressor Awareness}, or engaging the root stressor rather than surface symptoms. We also compute the Speech Response Rate ($RR$), which measures the fraction of turns in which the agent actively speaks instead of defaulting to silence.

\paragraph{Full-Duplex Behavior Evaluation.}
We evaluate all models on \textbf{Full-Duplex-Bench}~\cite{lin2025full}, which covers pause handling, backchanneling, turn-taking, and interruption management. This evaluation tests whether our therapeutic enhancements compromise the model's streaming agility.

\section{Results and Analysis}
\label{sec:results}
\subsection{Counseling Quality Assessment}
\label{subsec:counseling_quality}
Table~\ref{table_main_results} reports the GPT-4.1 judge scores on ESConv and ESCoT. CounselPlex achieves the highest score on all five dimensions for both datasets. It outperforms the fine-tuned baseline, PersonaPlex\textsuperscript{\textdagger}, by $+0.34$ Avg5 on ESConv and $+0.17$ on ESCoT.

The absolute scores remain in the 2--3 range, even for the ground-truth (GT) responses ($2.86$ on ESConv and $3.47$ on ESCoT). GT provides a practical but imperfect upper bound. The gap between CounselPlex and GT is wider on ESCoT ($-0.66$) than on ESConv ($-0.38$). The higher ESCoT GT score may partly reflect dataset construction: ESCoT responses are ChatGPT-generated, whereas ESConv contains natural crowdworker conversations. The difference may also reflect known biases in LLM-based evaluation~\cite{zheng2023judging}.

\begin{table}[t]
\centering
\caption{Cross-judge validation. Avg5 (silence-as-1) under two judge families on the combined ESConv + ESCoT test set ($535$ conversations). Both judges yield the identical ranking.}
\label{tab:crossjudge_main}
\small
\resizebox{0.65\linewidth}{!}{%
\begin{tabular}{lcc}
\toprule
\textbf{Model} & \textbf{GPT-4.1} & \textbf{Gemini-2.5} \\
\midrule
\cellcolor[HTML]{F0F0F0}\textbf{CounselPlex} & \cellcolor[HTML]{F0F0F0}\textbf{2.692} & \cellcolor[HTML]{F0F0F0}\textbf{2.926} \\
PersonaPlex$^{\dagger}$ & 2.461 & 2.691 \\
PersonaPlex & 2.278 & 2.595 \\
Moshi & 2.262 & 2.526 \\
\bottomrule
\end{tabular}
}
\end{table}

\paragraph{Robustness across judges.}
GPT-4.1 serves as both the stressor labeler and the judge, which could introduce evaluation bias. We therefore re-score all $535$ test conversations with Gemini-2.5-Flash, a model from a different judge family, using the same rubric. Table~\ref{tab:crossjudge_main} shows that Gemini reproduces the exact model ranking. CounselPlex remains first and leads PersonaPlex$^\dagger$ by the same margin ($+0.23$). The judges' per-response Avg5 scores have a Pearson correlation of approximately $0.73$ for CounselPlex, with similar correlations for the baselines (Appendix~\ref{appendix:cross_judge}).

\begin{table}[t]
\centering
\caption{Response-rate robustness on the combined ESConv and ESCoT test sets. Engaged-only Avg5 averages over each model's answered turns. Matched Avg5 uses only the $3{,}472$ turns answered by all four models.}
\label{tab:response_robustness}
\small
\resizebox{0.7\linewidth}{!}{%
\begin{tabular}{lcc}
\toprule
\textbf{Model} & \textbf{Engaged-only} & \textbf{Matched} \\
\midrule
\textbf{CounselPlex} & \textbf{2.804} & \textbf{2.812} \\
PersonaPlex (ZS) & 2.745 & 2.796 \\
Moshi & 2.590 & 2.656 \\
PersonaPlex$^\dagger$ (FT) & 2.544 & 2.537 \\
\bottomrule
\end{tabular}
}
\end{table}

\paragraph{Robustness to response rate.}
Our primary Avg5 assigns a score of 1 to silence, so unanswered turns are penalized. Table~\ref{tab:response_robustness} reports two additional views. Engaged-only Avg5 averages only the turns answered by each model, so the evaluated turns can differ across models. Matched Avg5 uses the same $3{,}472$ turns answered by all four models, so all models are compared on identical examples. CounselPlex ranks first in both views. On the matched turns, which are easier for the low-response-rate baselines (Appendix~\ref{appendix:rr_norm}), zero-shot PersonaPlex is statistically indistinguishable from CounselPlex ($2.796$ vs.\ $2.812$; $p{=}0.49$), whereas across all test turns the gap is larger ($2.278$ vs.\ $2.692$; Table~\ref{table_main_results}). CounselPlex also remains ahead of the fine-tuned PersonaPlex$^\dagger$ on the matched turns ($2.812$ vs.\ $2.537$). Its overall lead thus mainly reflects maintaining quality while answering more turns.

\paragraph{Human preference.}
In a pairwise A/B study conducted through the Prolific platform, five raters each evaluated $20$ randomly sampled response pairs. CounselPlex was preferred over PersonaPlex$^\dagger$ in $61.2\%$ of non-tie judgments. The full protocol, statistics, and per-rater results appear in Appendix~\ref{appendix:human_eval}.

\subsection{Duplex Interaction Trade-offs}
\label{subsec:duplex_behavior}
Table~\ref{tab:full_duplex_bench} shows that CounselPlex has a conservative interaction profile. It has the lowest Take-Over Rates during both synthetic and naturalistic Candor pauses ($0.066$ and $0.213$), meaning that it rarely takes the floor during within-turn pauses. At the same time, it responds in $99.2\%$ of the smooth turn-taking examples (TOR $0.992$), although its average latency is $0.988$\,s.

This interaction profile is accompanied by slower responses. CounselPlex responds after nearly all user interruptions (TOR $0.990$), but its latency is $1.056$\,s and its interruption-response score is $3.071$, compared with $0.400$\,s and $4.210$ for zero-shot PersonaPlex. Relative to the in-domain fine-tuned baseline PersonaPlex$^\dagger$, the full CounselPlex system is $0.068$\,s slower for smooth turn taking and $0.085$\,s slower after interruption. These small differences show that the two fine-tuned systems have similar latency, but do not isolate the source of the delay.

CounselPlex produces no detected backchannels (Freq $0.000$; JSD $1.000$), which is consistent with the absence of overlapping backchannel supervision in its turn-structured training data. Adding backchannel cues increases the measured frequency, but also raises Take-Over Rates in both the backchannel and pause settings (Appendix~\ref{appendix:backchannel}). Overall, CounselPlex responds in nearly all smooth-turn and interruption examples, but it is slower and produces less overlap, including no detected backchannels. Recovering natural backchannels without increasing unwanted takeovers remains important future work.

\subsection{Ablation Studies}
\label{subsec:ablation}
\begin{table}[t]
\caption{Ablation study of CounselPlex's core components on ESConv and ESCoT. \textbf{Base + In-domain Data}: backbone fine-tuned on our data without stressor labels. \textbf{w/o RPSA}: trained without RoPE Position Shift Augmentation. \textbf{w/o Inj.}: stressor is generated but not injected during listening.}
\centering
\resizebox{\linewidth}{!}{
\begin{tabular}{lccccccc}
\hline
Model & RR & Emp & Coh & Ctx & Strat & Strs & Avg \\ \hline
Baseline (PersonaPlex) & 0.733 & 2.144 & 2.410 & 2.323 & 2.179 & 2.336 & 2.278 \\
Base + In-domain Data & \textbf{0.946} & 2.331 & 2.547 & 2.543 & 2.354 & 2.532 & 2.461 \\
CounselPlex w/o RPSA & 0.882 & 2.103 & 2.246 & 2.285 & 2.109 & 2.263 & 2.201 \\
CounselPlex w/o Inj. & 0.901 & 2.218 & 2.392 & 2.411 & 2.228 & 2.395 & 2.329 \\
\cellcolor[HTML]{F0F0F0}CounselPlex & \cellcolor[HTML]{F0F0F0}0.938 & \cellcolor[HTML]{F0F0F0}\textbf{2.548} & \cellcolor[HTML]{F0F0F0}\textbf{2.797} & \cellcolor[HTML]{F0F0F0}\textbf{2.772} & \cellcolor[HTML]{F0F0F0}\textbf{2.577} & \cellcolor[HTML]{F0F0F0}\textbf{2.766}  & \cellcolor[HTML]{F0F0F0}\textbf{2.692}\\ \hline
\end{tabular}
}
\label{table:ablation}
\end{table}

Table~\ref{table:ablation} indicates that both RPSA and stressor injection contribute in our setup. Removing either component lowers Response Rate and counseling quality below the in-domain fine-tuned baseline. Generating a think trace without reinjecting it is therefore insufficient in this configuration. The complete \textbf{CounselPlex} model achieves the best scores across all counseling metrics.

\subsection{Analysis of Stressor Reasoning}
\label{subsec:stressor_analysis}
Section~\ref{subsec:ablation} suggests that the stressor-related components contribute to CounselPlex's gains. We next examine their behavior during inference: how often and how accurately the model generates a stressor state, and how well that state persists as a conversation progresses.

\paragraph{Stressor reasoning is used widely and accurately.}
CounselPlex generates a new stressor state on $11.3\%$ of engaged turns. At least one new state appears in $83.2\%$ of conversations, with an average of $1.24$ new states per conversation. This mean is close to the training density of $1.19$ per conversation (Section~\ref{subsec:data}). A GPT-4o-mini judge finds that $92.5\%$ of the $666$ generated stressor states correctly capture the seeker's underlying situation.

\begin{table}[t]
\centering
\caption{Counseling quality (Avg5) and stressor verbalization rate by stressor context. None (n=366): no stressor in context; New (n=521): newly introduced this turn; Reused (n=4,780): earlier stressor still active. Verbalization rate is undefined for None.}
\label{table:verbalization}
\resizebox{0.9\columnwidth}{!}{
\begin{tabular}{lccccc}
\toprule
& \multicolumn{3}{c}{\textbf{Counseling Quality (Avg5)}} & \multicolumn{2}{c}{\textbf{Stressor Verbalization}} \\
\cmidrule(lr){2-4} \cmidrule(lr){5-6}
\textbf{Model} & None & New & Reused & New & Reused \\
\midrule
Moshi        & 2.102 & 2.895 & 2.259 & 66.8\% & 28.0\% \\
PersonaPlex  & 2.320 & 2.645 & 2.273 & 49.7\% & 23.8\% \\
\rowcolor[HTML]{F0F0F0}
CounselPlex  & \textbf{2.734} & \textbf{3.147} & \textbf{2.800} & \textbf{61.8\%} & \textbf{36.2\%} \\
\hdashline
GT (reference) & 2.940 & 3.256 & 3.259 & 85.0\% & 45.9\% \\
\bottomrule
\end{tabular}
}
\end{table}

\paragraph{CounselPlex narrows the gap to reference responses in sustaining stressor grounding.} Table~\ref{table:verbalization} reports counseling quality (Avg5) and \emph{stressor verbalization} across different stressor contexts. Stressor verbalization is the rate at which a response meaningfully engages with the active reference stressor by acknowledging, exploring, or addressing it. Using the annotations from Section~\ref{subsec:data}, we divide engaged turns into three groups: \emph{none}, with no stressor in context ($n{=}366$); \emph{new}, with a label introduced in the current turn ($n{=}521$); and \emph{reused}, with an earlier label that remains active ($n{=}4{,}780$). GPT-4o-mini makes this binary judgment using the reference stressor active at each turn (Appendix~\ref{appendix:verbalization}).

Two findings stand out. First, every model has both lower quality and less stressor verbalization on \emph{reused} turns than on \emph{new} turns. GT responses also verbalize the stressor less often on \emph{reused} turns, but they maintain their quality. This pattern suggests that the reference responses use the stressor as a lens rather than simply restating it. CounselPlex shows the same blindness as the other models, but to a lesser degree: on \emph{reused} turns, its Avg5 gap to GT is $0.46$, compared with $0.99$--$1.00$ for the baselines.

Second, CounselPlex approaches GT verbalization, especially on \emph{reused} turns (36.2\% vs.\ GT 45.9\%). The baselines fall further behind. Thus, CounselPlex improves not only response quality but also the way its language engages with the stressor. Appendix~\ref{appendix:multi_stressor} further analyzes conversations with multiple stressors.

\begin{figure}[t]
\centering
\includegraphics[width=0.75\columnwidth]{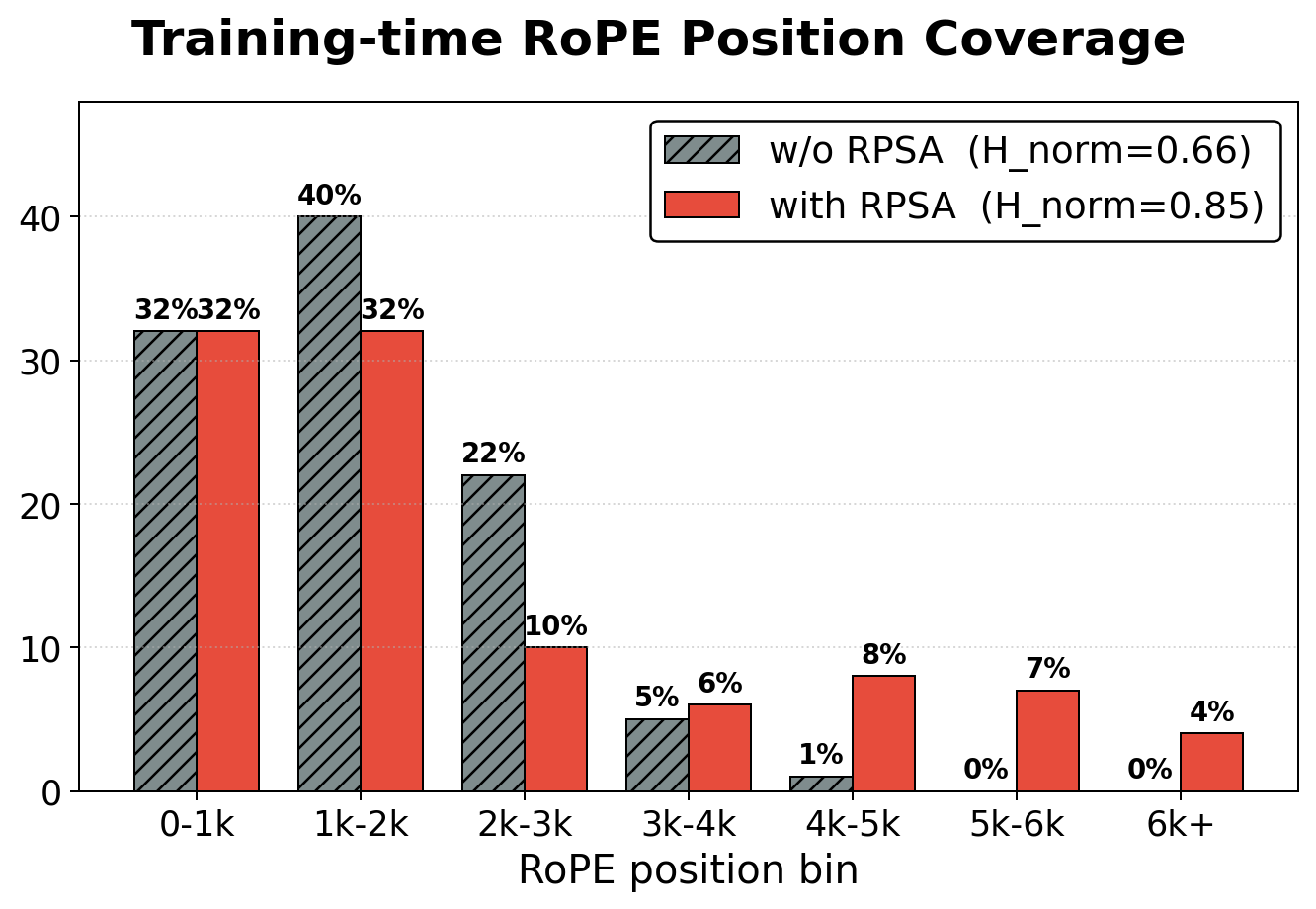}
\caption{\textbf{Histogram of RoPE positions seen during training.} RPSA broadens positional coverage (normalized entropy $H_{\text{norm}}: 0.66\rightarrow0.85$) and exposes the model to higher indices that the baseline never reaches.}
\label{fig:rpe_coverage}
\end{figure}

\begin{figure}[t]
\centering
\includegraphics[width=0.9\columnwidth]{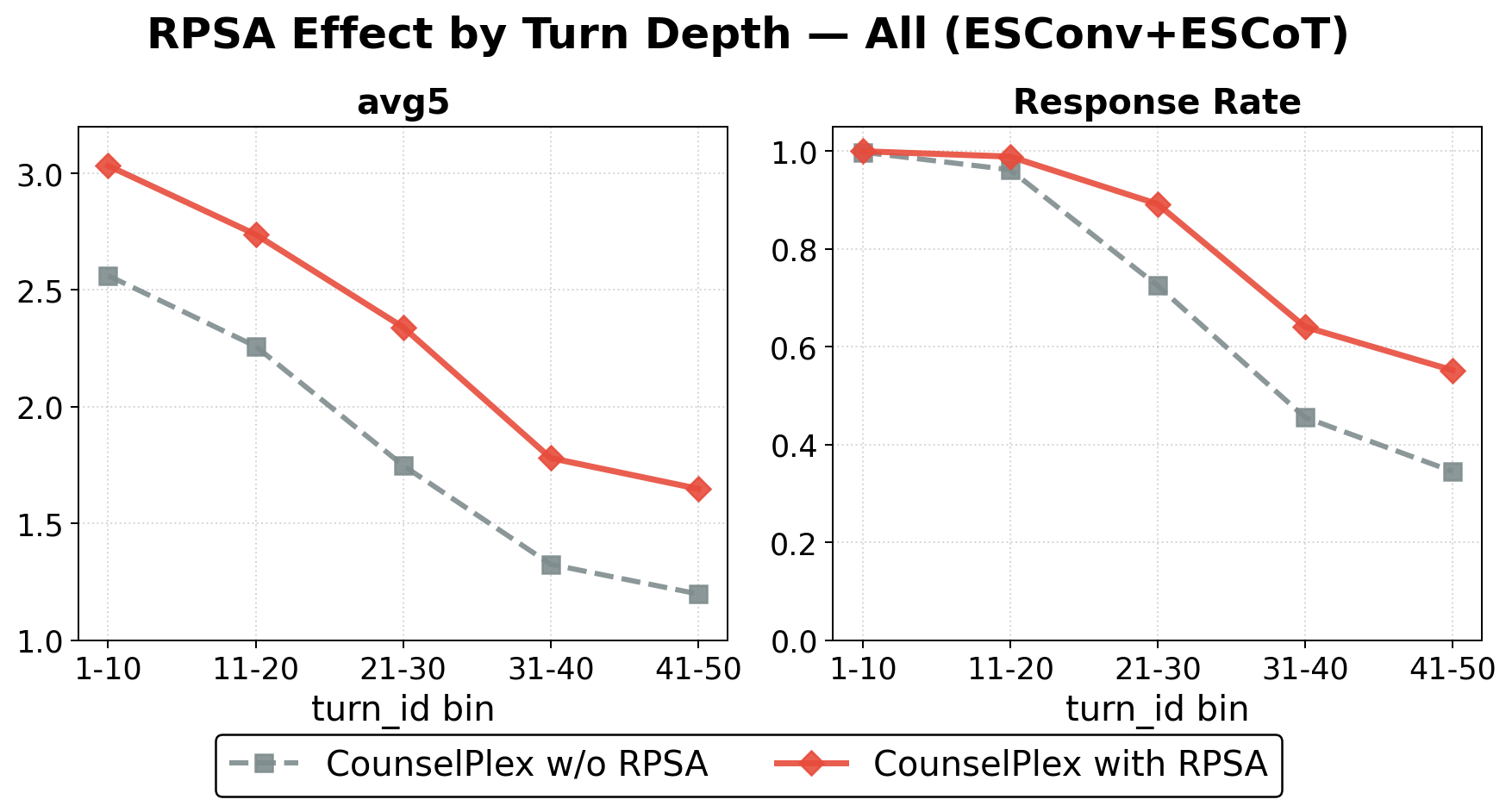}
\caption{\textbf{RPSA effect by turn depth.} Counseling quality (Avg5) and Response Rate on the combined ESConv and ESCoT test sets. RPSA maintains substantially higher performance as conversations become longer. Bins 1--30 cover $87.1\%$ of test turns; bins 31+ are sparse.}
\label{fig:rpsa_turn_depth}
\end{figure}
\subsection{Long-Context Stability \& RPSA Effect}
\label{subsec:rpsa_effect}
Counseling dialogues vary widely in length. However, training samples are concentrated in the lower RoPE index range (Figure~\ref{fig:rpe_coverage}), so higher positions are under-trained. RPSA reduces this imbalance by randomly shifting the starting RoPE position of long-context samples. This shift exposes the model to a broader range of positions during training.

Without RPSA, both quality and Response Rate decline as turn depth increases (Figure~\ref{fig:rpsa_turn_depth}). RPSA maintains higher performance over the first 30 turns, which contain 87.1\% of test turns. A residual gap remains beyond turn 30.

\subsection{Partial Long-Range Grounding}
\label{sec:case_study}
\begin{table}[t]
\centering
\caption{Partial long-range grounding: CounselPlex returns to a family conflict after 16 turns but asks about previously disclosed details.}
\label{tab:main_case_study}
\small
\resizebox{0.95\linewidth}{!}{%
\begin{tabular}{@{}p{\columnwidth}@{}}
\toprule
\textbf{Injected Stressor State:} family conflict \\
\midrule
\textbf{Seeker (turn $t-16$, Initial Disclosure):} Well, my sister and I aren't speaking right now, and I feel really hurt and frustrated about it. It's been going on for months\ldots \\
\rowcolor[HTML]{F0F0F0}\textbf{$\hookrightarrow$ CounselPlex Internal Update:} \texttt{<think> family conflict </think>} \\
\textbf{Earlier disclosures (summary; turns $t-14$ and $t-6$):} A major disagreement escalated, and the sister later accused the seeker of not being there for her. \\
\midrule
\multicolumn{1}{c}{\textcolor{gray}{\textit{\ldots{} intervening turns; stressor state continuously injected \ldots}}} \\
\midrule
\textbf{Supporter (turn $t-1$):} Would it be helpful for you to try some relaxation techniques, like deep breathing or meditation? \\
\textbf{Seeker (current):} Maybe. I've never really tried anything like that before. \\
\midrule
\textbf{Moshi:} It might help to talk through some of your thoughts and feelings with someone who is trained to listen. \\
\textbf{PersonaPlex$^\dagger$:} There are many resources available that may be of help. Can you give me more information? \\
\rowcolor[HTML]{F0F0F0}\textbf{CounselPlex (Ours):} Can you tell me more about what's been going on that's caused this tension with your sister? Have you two been arguing a lot, or are there other factors at play here? \\
\textbf{GT (reference):} There are a lot of apps and resources available that can guide you through relaxation exercises, would you like me to send some recommendations your way? \\
\bottomrule
\end{tabular}
}
\end{table}

Figure~\ref{fig:stressor_blindness} shows a clear instance of stressor reconnection, whereas Table~\ref{tab:main_case_study} presents a harder long-context case. After the seeker discloses a conflict with their sister, the discussion shifts to coping and meditation. Sixteen turns later, CounselPlex reconnects its response to the family conflict, while the baseline responses remain generic. However, it asks about details that the seeker had already provided, and the reference response instead follows the immediate discussion of meditation. The example therefore shows topic-level stressor persistence, but not complete context recall or an unambiguously better response. Additional success and failure cases appear in Appendices~\ref{appendix:case_studies} and~\ref{appendix:failure_cases}.

\section{Discussion}
\label{sec:discussion}

\paragraph{Counseling gains beyond explicit stressor mentions.}
CounselPlex improves all five counseling dimensions over the in-domain PersonaPlex$^\dagger$ baseline on both ESConv and ESCoT. The improvement is therefore not confined to Stressor Awareness. Gemini-2.5-Flash reproduces the same model ranking, and a small human preference study favors CounselPlex in $61.2\%$ of non-tie comparisons. Together, these findings show that the observed gains extend beyond the explicitly targeted Stressor Awareness metric. However, the current evaluation does not isolate why each counseling dimension improves.

\paragraph{What the ablations establish.}
The ablations indicate that both stressor injection and RPSA contribute to the complete system. In our setup, generating a stressor trace without reinjecting it during later seeker speech is insufficient to improve over the in-domain baseline. The variant with RPSA also maintains higher Avg5 and Response Rate at later turn depths. These findings suggest that the complete model's gains cannot be explained by in-domain fine-tuning alone, but they do not establish that injection and RPSA address independent failure modes.

\paragraph{Duplex interaction trade-offs.}
CounselPlex has low Take-Over Rates during synthetic and naturalistic pauses while responding in $99.2\%$ of smooth-turn examples. However, it responds more slowly, has a lower interruption-response score than zero-shot PersonaPlex ($3.071$ vs.\ $4.210$), and produces no detected backchannels. Thus, reliable responses at clear turn boundaries coexist with delayed and less-overlapping behavior. Whether this trade-off is preferable in counseling requires domain-specific evaluation.

\paragraph{Behavior with multiple stressors.}
CounselPlex explicitly maintains one active stressor while the remaining dialogue history stays in context. Of the $80$ analyzed conversations containing a second stressor, $52$ ($65.0\%$) include at least one response that refers to a previously discussed stressor, either alone or together with the current stressor. A stricter subset of $38$ conversations ($47.5\%$) includes at least one response that refers to both the current and a previous stressor together. These observations show that a single active state does not always prevent reference to earlier dialogue information. Because the analysis is descriptive and has no controlled baseline, it does not establish that the active state functions as a priority mechanism.

\paragraph{Scope.}
Our experiments evaluate stressor grounding and interaction behavior, not clinical effectiveness or therapeutic outcomes. The speech is synthesized from text-based emotional-support datasets, and counseling quality is evaluated primarily from text-stream outputs using LLM judges. We therefore position CounselPlex as a counseling-oriented emotional-support model rather than a clinically validated counseling system.

\section{Conclusion}
\label{sec:conclusion}
We introduced \textbf{CounselPlex}, a full-duplex counseling framework for \emph{stressor blindness}. It generates a compressed stressor state during seeker silence and injects that state into \texttt{<PAD>} slots during active listening. In our evaluation, RPSA improves performance at longer turn depths. CounselPlex outperforms strong baselines across five counseling dimensions and narrows the stressor-verbalization gap to the reference responses. Full-Duplex-Bench shows low pause Take-Over Rates alongside slower responses and no detected backchannels. These results motivate further study of explicit, persistent stressor grounding in counseling-oriented conversational agents.

\bibliography{main}
\clearpage
\appendix
\section{Stressor Labeling Prompt and Examples}
\label{appendix:stress_labeling_prompt}

Table~\ref{table:stressor_labeling} presents the full prompt used to make GPT-4.1 track and extract the core stressor in each counseling dialogue.

The prompt imposes two constraints to keep state injection efficient. First, the model must distinguish concrete situations from temporary emotional symptoms and express each stressor as a 1--7 word noun phrase. Second, it cannot accumulate multiple stressors. If the seeker's main focus changes, the new stressor replaces the previous one.

Representative examples of stressor states extracted under these instructions include: \textit{covid job crisis}, \textit{emotionally abusive partner}, \textit{new parent adjustment}, \textit{father death}, \textit{job loss}, \textit{friend falling out}, \textit{career and personal life stagnation}, \textit{financial hardship}, and \textit{relationship difficulties}.

\begin{center}
\captionsetup{hypcap=false} 
\captionof{table}{Prompt template used with GPT-4.1 for extracting core stressor states from counseling dialogues (Appendix~\ref{appendix:stress_labeling_prompt}).}
\label{table:stressor_labeling}
\begin{tcolorbox}[
    breakable, 
    left=0.5mm, right=0.5mm, top=0.5mm, bottom=0.5mm, 
    fontupper=\tiny, 
    arc=0mm, boxrule=0.5pt,
    colback=white
]
\begin{lstlisting}[style=prompt]
Read the full counseling conversation. Output a sequence of STANDALONE
SNAPSHOTS marking each turn where the seeker's CURRENT core stressor 
changes.

[CRITICAL] — REPLACE, DO NOT ACCUMULATE:
Each snapshot is independent. It captures ONLY what is the MOST IMPORTANT
core stressor at THAT turn. 
Do NOT carry forward earlier stressors with "+" or "&".
Each new entry REPLACES the previous one entirely.

A core stressor is an actionable underlying SITUATION (not symptom).

------------ WHAT COUNTS / WHAT DOESN'T ------------

[O] Stressors: concrete events / circumstances / ongoing burdens
- Life events: "friend death", "job loss", "breakup", "abusive parent"
- Chronic health conditions: "migraines", "chronic pain", "diabetes
     diagnosis", "long-term illness", "physical disability",
     "chronic insomnia" (when sleep IS the primary issue, not a symptom)
- Long-standing mental health diagnoses with treatment context:
     "managing bipolar diagnosis", "PTSD from past event", "OCD diagnosis"
     (label the DIAGNOSIS as a stressor, not the symptom)

[X] NOT stressors (do not include):
- Pure emotional symptoms: "sleepless", "sad", "anxious", "lonely",
     "hopeless", "self-doubt"  <- without a concrete underlying situation
- Coping behaviors: "trying meditation", "joined gym", "going to therapy"
- Trivial: "weekend plans", "hobby preferences"

[WARNING] Important distinction:
- "Feeling depressed" alone -> [X] symptom (no stressor)
- "Depression diagnosis affecting daily life"
  -> [O] chronic mental health stressor
- "Hopeless feeling" alone -> [X] symptom
- "Migraines making it hard to function" -> [O] chronic health stressor
- "Sleepless from grief" -> [X] symptom of an underlying loss
- "Chronic insomnia, can't sleep most nights" (no other cause) 
  -> [O] sleep disorder is THE issue

------------ SNAPSHOT STYLE ------------

KEYWORD TAGS, NOT SENTENCES. 1-7 words HARD CAP — never more.
- Noun-phrase or subject+state form
- NO articles, NO prepositions, NO filler verbs
- NO "+" or "&" joining of multiple stressors — pick the ONE most pressing
  at this moment, or use a single umbrella term if two are inseparable

GOOD: "covid friend death", "appearance anxiety", "job loss", "breakup",
      "covid household stress", "marital infidelity", "caregiving burden"
BAD:  "lost childhood best friend to covid 19" (sentence, too long)
BAD:  "job loss + breakup" (accumulation, not allowed in v15)
BAD:  "sleepless from grief" (symptom, not stressor)

------------ WHEN TO ADD A NEW SNAPSHOT ------------

Add a new entry ONLY when the MAIN stressor genuinely shifts:
- A new, distinct stressor takes over the seeker's current focus
- A previously vague concern becomes specific enough to relabel

Do NOT add entries for:
- Symptom elaboration of the same stressor
- Coping discussion
- Restating the same stressor in different words

Most convs have just 1 snapshot (the main stressor, introduced once).
2-3 snapshots only if the conv genuinely shifts focus to a different 
stressor.

------------ NO STRESSOR ------------

If seeker only expresses emotions / coping with no underlying situation,
return {"updates": []}.

------------ EXAMPLES ------------

Example A (single stressor, never changes):
  Conv: turn 3 seeker reveals friend died of covid; rest of conv discusses
  sleepless nights and memories of the friend.
  Output:
    {"updates": [
        {"turn": 3, "snapshot": "covid friend death"}
    ]}

Example B (focus genuinely shifts):
  Conv: turn 1 seeker mentions job loss; turn 12 they reveal a separate
  relationship breakup that's now their main concern.
  Output:
    {"updates": [
        {"turn": 1, "snapshot": "job loss"},
        {"turn": 12, "snapshot": "breakup"}
    ]}
  [WARNING] Turn 12 REPLACES, does NOT accumulate.

------------ OUTPUT FORMAT ------------

Strict JSON:
  {"updates": [
      {"turn": <seeker turn id>, "snapshot": "<1-7 word standalone tag>"},
      ...
  ]}

------------ CONVERSATION ------------

problem_type: {problem_type}
emotion_type: {emotion_type}

{conv_text}
\end{lstlisting}
\end{tcolorbox}
\end{center}

\begin{table*}[t]
\centering
\caption{Strategy distribution and diversity under the eight-class ESConv taxonomy. Strategy shares are percentages. Higher entropy indicates greater diversity; lower Jensen--Shannon divergence (JSD) indicates a distribution closer to the reference responses. Bold marks the best model result in each summary row.}
\label{tab:strategy_distribution}
\small
\setlength{\tabcolsep}{5pt}
\renewcommand{\arraystretch}{1.05}
\begin{tabular*}{\textwidth}{@{\extracolsep{\fill}}lcccc@{\hspace{8pt}}c}
\toprule
\textbf{Strategy} & \textbf{Moshi} & \textbf{PP (ZS)} & \textbf{PP$\dagger$ (FT)} & \textbf{CounselPlex} & \textbf{Reference} \\
\midrule
Question & 10.4 & 16.7 & 32.9 & 37.3 & 21.9 \\
Restatement & 1.9 & 4.1 & 5.9 & 8.2 & 3.3 \\
Reflection of Feelings & 2.7 & 5.2 & 4.5 & 6.7 & 17.0 \\
Self-disclosure & 2.5 & 6.0 & 7.1 & 6.2 & 5.6 \\
Affirmation & 54.0 & 37.1 & 23.8 & 21.4 & 17.3 \\
Providing Suggestions & 18.6 & 23.5 & 17.5 & 16.5 & 23.9 \\
Information & 2.2 & 1.7 & 2.2 & 1.3 & 4.4 \\
Others & 7.7 & 5.8 & 6.1 & 2.5 & 6.5 \\
\midrule
Shannon $H$ (bits) $\uparrow$ & 2.061 & 2.442 & \textbf{2.543} & 2.453 & 2.698 \\
$H_{\text{norm}}$ $\uparrow$ & 0.687 & 0.814 & \textbf{0.848} & 0.818 & 0.899 \\
JSD to GT $\downarrow$ & 0.140 & 0.058 & \textbf{0.049} & 0.059 & --- \\
\midrule
Top-1 Strategy & Affirm. & Affirm. & Quest. & Quest. & Sugg. \\
Top-1 Share (\%) & 54.0 & 37.1 & 32.9 & 37.3 & 23.9 \\
Top-3 Coverage (\%) & 82.9 & 77.3 & 74.1 & 75.2 & 63.1 \\
\bottomrule
\end{tabular*}
\end{table*}


\section{Spoken-Corpus Construction}
\label{appendix:data_construction}

We use ESConv~\citep{liu2021towards} and ESCoT~\citep{zhang2024escot} because both contain multi-turn, strategy-oriented emotional-support dialogues. They therefore provide the long conversational histories needed to study stressor retention, while their supporter responses enable direct comparison with the models' responses. We retain the source datasets' train, validation, and test assignments, yielding $2{,}107$, $367$, and $535$ conversations, respectively.

Because both source datasets are text-only, we synthesize every seeker and supporter utterance with CosyVoice3~\citep{du2025cosyvoice}. A fixed male voice is used for the seeker and a fixed female voice for the supporter in every conversation. The resulting corpus contains $80{,}034$ utterance-level WAV files at $24$\,kHz, totaling $158.3$ hours across $3{,}009$ conversations. This fixed two-voice design keeps the acoustic roles consistent during training, but it does not represent the speaker, accent, or prosodic diversity of real counseling speech.

GPT-4.1 processes each dialogue chronologically using the prompt in Table~\ref{table:stressor_labeling}. It emits a new label only when the seeker's main situational stressor is introduced or changes; otherwise, the previous label remains active. We leave $127$ conversations ($4.22\%$) unlabeled because the dialogue does not state an external stressor.

\paragraph{Release scope.}
We do not currently distribute the derived WAV files, stressor annotations, or full preprocessing pipeline. The experiments require role-aligned speech and timing information that is absent from the publicly available text corpora, so the text alone is not an equivalent training input for a full-duplex speech model. Any future release remains contingent on the redistribution terms of the source data and voice assets.



\section{Strategy Selection and Diversity Analysis}
\label{appendix:strategy_diversity}

Table~\ref{tab:strategy_distribution} compares supporter strategies under the 8-class ESConv taxonomy. Moshi relies heavily on \textit{Affirmation} (54.0\%) and has the lowest diversity ($H_{\text{norm}} = 0.687$). In CounselPlex, \textit{Question} is the most frequent strategy (37.3\%). CounselPlex is more diverse than the zero-shot baselines, but the reference-response distribution (GT) remains the most balanced ($H_{\text{norm}} = 0.899$). GT also assigns less mass to its three most frequent strategies (63.1\%) than any model, indicating broader use of the less common strategy classes. These distributions are descriptive; they do not isolate stressor reasoning as the sole cause of the differences between models.

\section{Latency Breakdown of Stressor Reasoning}
\label{appendix:latency}

\paragraph{Speech-onset latency.}
We analyze the complete held-out test split: $535$ conversations
($194$ from ESConv and $341$ from ESCoT), comprising $6{,}278$
seeker--supporter turns. All measurements use the final CounselPlex
checkpoint with inference seed $42$, so the audio and stressor-state
labels come from the same decoding run. We partition turns according
to that run's trace log: \emph{new-update} turns generate a new
stressor trace, \emph{reuse} turns retain an existing stressor without
generating a new trace, and \emph{no-active-stressor} turns have no
stored stressor. After each seeker turn, CounselPlex waits for
$0.5$\,s of silence before deciding whether a new stressor is needed.
If an update is needed, the model generates the trace after this
check. We measure latency from the end of seeker speech to the start
of CounselPlex speech. Thus, this measurement already includes the
$0.5$\,s silence check.

Of the $6{,}278$ test turns, Silero detects response onset for
$5{,}627$ turns. Whisper timestamps are available for $5{,}626$ of
these same turns. With Silero, mean onset
latency is $1.668$\,s on new-update turns and $1.320$\,s on turns with
no active stressor. The difference is $0.348$\,s. With Whisper, the
corresponding values are $1.192$\,s and $0.828$\,s, a difference of
$0.364$\,s. On turns that reuse an existing stressor, mean onset
latency is $1.303$\,s with Silero and $0.826$\,s with Whisper. These
reuse values are close to the values for turns with no active stressor
because no new trace is generated.

These comparisons are observational: the turn types occur naturally
and are not randomly assigned. Therefore, the differences should not
be interpreted as causal effects. This protocol also differs from
Full-Duplex-Bench (Table~\ref{tab:full_duplex_bench}), so their
absolute latency values should not be compared directly.

\paragraph{Length of generated stressors.}
We analyze the $666$ new traces reported in
Section~\ref{subsec:stressor_analysis}. Each trace contains $2.62$
words on average. With SentencePiece tokenization, the stressor text
contains $2.98$ tokens on average. The opening and closing
\texttt{<think>} tokens increase the total to $4.98$ tokens.

The text stream runs at $12.5$ frames per second and carries one text
token per frame. Thus, one trace occupies $0.398$\,s (about
$0.40$\,s) of the text stream on average. This duration does not
include the earlier $0.5$\,s silence check. However, the end-to-end
latency reported above already includes the check and all later
processing. At $1.24$ updates per conversation, the traces occupy
$0.496$\,s (about $0.50$\,s) in total. This value measures internal
text-stream time.

\section{Case Studies}
\label{appendix:case_studies}

We complement the quantitative results in Section~\ref{sec:results} with five qualitative case studies. In each case, the seeker discloses a stressor early in the conversation. CounselPlex compresses it into a short \texttt{<think>} trace and injects that trace in later turns. The cases cover workplace pressure, relationship conflict, appearance anxiety, caregiving burden, and noisy living conditions. We compare CounselPlex with Moshi, PersonaPlex$^\dagger$, and the dataset reference response (GT) to examine how each response uses the earlier stressor after the dialogue shifts toward symptoms or coping.

Three patterns recur across the cases. First, CounselPlex names the underlying tension instead of immediately offering a solution or generic validation (Cases~1 and~3). Second, it can follow the seeker's emotional direction rather than remaining fixed on the past stressor. In Case~2, for example, it shifts from the unresolved relationship to what the seeker now wants from a partner. Third, it explicitly links current feelings to the earlier stressor, even when the latest utterance concerns a related but different feeling (Case~5).

\begin{table}[tbp]
\centering
\small
\caption{Case 1: Workplace deadline pressure.}
\label{tab:case_studies_part1}
\setlength{\tabcolsep}{4pt}
\renewcommand{\arraystretch}{1.02}
\begin{tabular}{p{\dimexpr\columnwidth-2\tabcolsep\relax}}
\toprule
\rowcolor{casebluebg}\textbf{Injected stressor:} workplace project deadline pressure \\
\textbf{Seeker (turn $t-11$, original disclosure):} I have a really tight deadline for a major project at work, and I just can't seem to focus or get anything done. \\
\textcolor{caseblue}{\textbf{$\hookrightarrow$ CounselPlex internal update:}} \texttt{<think> workplace project deadline pressure </think>} \\
\midrule
\multicolumn{1}{c}{\textcolor{casemuted}{\itshape \ldots{} six intervening turns; stressor state retained \ldots{}}} \\
\midrule
\textbf{Supporter (turn $t{-}2$):} Maybe practicing what you want to say beforehand could help you feel more prepared. \\
\textbf{Seeker (turn $t{-}1$):} Yeah, that might help. \\
\textbf{Supporter (turn $t{-}1$):} You could also try expressing your concerns by suggesting alternatives or improvements rather than complaining. \\
\textbf{Seeker (current):} \textit{That makes sense, but I don't want to come across as complaining or difficult.} \\
\midrule
\textbf{Moshi:} It is essential for both of you as an employee and employer to maintain a healthy work environment. \\
\textbf{PersonaPlex:} Of course, you can frame it as a problem you're trying to solve together. \\
\textbf{PersonaPlex\textsuperscript{\textdagger}:} Have you talked to any of your other coworkers about it? \\
\rowcolor{casegreenbg}\textcolor{casegreen}{\textbf{CounselPlex (ours):}} There is a fine line between being someone who stands up for themselves and feels comfortable in their workload, and feeling like you're causing a conflict. It sounds like you're trying to balance that situation in your workplace. \\
\textbf{Reference (GT):} I understand your concerns, but it's important to remember that your boss may not even be aware of the impact their behavior is having on you. \\
\bottomrule
\end{tabular}
\end{table}

\begin{table}[tbp]
\centering
\small
\caption{Case 2: Relationship conflict.}
\label{tab:case_study_2}
\setlength{\tabcolsep}{4pt}
\renewcommand{\arraystretch}{1.02}
\begin{tabular}{p{\dimexpr\columnwidth-2\tabcolsep\relax}}
\toprule
\rowcolor{casebluebg}\textbf{Injected stressor:} relationship conflict \\
\textbf{Seeker (turn $t-12$, original disclosure):} Sorry about that. This is about my relationship. \\
\textcolor{caseblue}{\textbf{$\hookrightarrow$ CounselPlex internal update:}} \texttt{<think> relationship conflict </think>} \\
\midrule
\multicolumn{1}{c}{\textcolor{casemuted}{\itshape \ldots{} six intervening turns; stressor state retained \ldots{}}} \\
\midrule
\textbf{Supporter (turn $t{-}2$):} Have you had a conversation with her about her actions and how they have made you feel? \\
\textbf{Seeker (turn $t{-}1$):} She knew pretty well how I felt. I don't want to have anything to do with her again. \\
\textbf{Supporter (turn $t{-}1$):} Do you feel that her actions and result can be remedied? Or is this such a fundamental break that you cannot reconcile? \\
\textbf{Seeker (current):} \textit{I just want to move on with my life.} \\
\midrule
\textbf{Moshi:} It's important not to let the past rule your life too much. \\
\textbf{PersonaPlex\textsuperscript{\textdagger}:} You will get back on your feet. I understand you now. I didn't want going on with my ex husband but I did it for the kids. \\
\rowcolor{casegreenbg}\textcolor{casegreen}{\textbf{CounselPlex (ours):}} I wonder if you could tell me more about what you need from a partner? What do you feel you deserve in a relationship that will last? \\
\textbf{Reference (GT):} Understandable. I would feel similar, in truth. \\
\bottomrule
\end{tabular}
\end{table}

\begin{table}[tbp]
\centering
\small
\caption{Case 3: Appearance anxiety.}
\label{tab:case_studies_part2}
\setlength{\tabcolsep}{4pt}
\renewcommand{\arraystretch}{1.02}
\begin{tabular}{p{\dimexpr\columnwidth-2\tabcolsep\relax}}
\toprule
\rowcolor{casebluebg}\textbf{Injected stressor:} appearance anxiety \\
\textbf{Seeker (turn $t-11$, original disclosure):} I just can't stop thinking about how I should look and how I fall short. \\
\textcolor{caseblue}{\textbf{$\hookrightarrow$ CounselPlex internal update:}} \texttt{<think> appearance anxiety </think>} \\
\midrule
\multicolumn{1}{c}{\textcolor{casemuted}{\itshape \ldots{} six intervening turns; stressor state retained \ldots{}}} \\
\midrule
\textbf{Supporter (turn $t{-}2$):} Have you ever tried practicing self-care activities like meditation or yoga? \\
\textbf{Seeker (turn $t{-}1$):} No, I haven't. \\
\textbf{Supporter (turn $t{-}1$):} Research has shown that engaging in activities like yoga, meditation, or deep breathing exercises can have a calming effect on the mind and body and help reduce stress and anxiety. \\
\textbf{Seeker (current):} \textit{That sounds interesting.} \\
\midrule
\textbf{Moshi:} There's actually a great place in Los Angeles where creatives gather to share their ideas... \\
\textbf{PersonaPlex\textsuperscript{\textdagger}:} How do you think you could work on becoming more mindful of your thoughts and behaviors? \\
\rowcolor{casegreenbg}\textcolor{casegreen}{\textbf{CounselPlex (ours):}} You're struggling with appearance anxiety but you're willing to take steps toward feeling more confident. \\
\textbf{Reference (GT):} Why don't you try taking a yoga class or downloading a meditation app? Even if it feels strange at first, it might help you feel more grounded in your own body. \\
\bottomrule
\end{tabular}
\end{table}

\begin{table}[tbp]
\centering
\small
\caption{Case 4: Caregiving burden.}
\label{tab:case_study_4}
\setlength{\tabcolsep}{4pt}
\renewcommand{\arraystretch}{1.02}
\begin{tabular}{p{\dimexpr\columnwidth-2\tabcolsep\relax}}
\toprule
\rowcolor{casebluebg}\textbf{Injected stressor:} caregiving burden \\
\textbf{Seeker (turn $t-13$, original disclosure):} Honestly, I'm feeling pretty exhausted. I'm caring for my loved one with a chronic illness, and it's really taking a toll on my own physical and mental health. \\
\textcolor{caseblue}{\textbf{$\hookrightarrow$ CounselPlex internal update:}} \texttt{<think> caregiving burden </think>} \\
\midrule
\multicolumn{1}{c}{\textcolor{casemuted}{\itshape \ldots{} six intervening turns; stressor state retained \ldots{}}} \\
\midrule
\textbf{Supporter (turn $t{-}2$):} It sounds like you're feeling hesitant about getting additional support due to financial concerns. \\
\textbf{Seeker (turn $t{-}1$):} Yes, that's definitely a factor. \\
\textbf{Supporter (turn $t{-}1$):} Did you know that there are programs and resources available to help caregivers with financial assistance? \\
\textbf{Seeker (current):} \textit{Really? I had no idea.} \\
\midrule
\textbf{Moshi:} \textit{(No speech generated)} \\
\textbf{PersonaPlex\textsuperscript{\textdagger}:} You've been doing such a great job, and I am in awe of your resilience and care... \\
\rowcolor{casegreenbg}\textcolor{casegreen}{\textbf{CounselPlex (ours):}} I want you to know that I'm here for you. You're taking a lot of responsibility on yourself, but you're not alone in this. There are organizations and support groups out there for people in your situation. \\
\textbf{Reference (GT):} It might be worth looking into those resources to see what's available to help alleviate some of the financial burden. \\
\bottomrule
\end{tabular}
\end{table}

\begin{table}[tbp]
\centering
\small
\caption{Case 5: Noisy neighbors.}
\label{tab:case_studies_part3}
\setlength{\tabcolsep}{4pt}
\renewcommand{\arraystretch}{1.02}
\begin{tabular}{p{\dimexpr\columnwidth-2\tabcolsep\relax}}
\toprule
\rowcolor{casebluebg}\textbf{Injected stressor:} noisy neighbors \\
\textbf{Seeker (turn $t-27$, original disclosure):} Honestly, it's been really tough. I live in a small apartment and my neighbors are incredibly noisy. I can't seem to get a decent night's sleep and it's affecting my work. \\
\textcolor{caseblue}{\textbf{$\hookrightarrow$ CounselPlex internal update:}} \texttt{<think> noisy neighbors </think>} \\
\midrule
\multicolumn{1}{c}{\textcolor{casemuted}{\itshape \ldots{} six intervening turns; stressor state retained \ldots{}}} \\
\midrule
\textbf{Supporter (turn $t{-}2$):} Have you thought about trying different types of noise-cancelling headphones or maybe even a white noise machine to help drown out the noise? \\
\textbf{Seeker (turn $t{-}1$):} No, I hadn't thought of that. I'll look into it. \\
\textbf{Supporter (turn $t{-}1$):} Have you spoken to any friends or family members about what you're going through? \\
\textbf{Seeker (current):} \textit{Not really, I feel like I'm bothering them with all my problems.} \\
\midrule
\textbf{Moshi:} It can be overwhelming at times but remember that they care about you and want to see you happy too. \\
\textbf{PersonaPlex:} I hear you. It's tough. We're all dealing with our own stuff, right? \\
\textbf{PersonaPlex\textsuperscript{\textdagger}:} They're external bodies, they have no idea your problems. \\
\rowcolor{casegreenbg}\textcolor{casegreen}{\textbf{CounselPlex (ours):}} It sounds like you're feeling like you're bothering them. Can you tell me more? I think this might relate to what you're feeling. \\
\textbf{Reference (GT):} You're not a burden for reaching out for help. It takes courage to share your struggles with others. \\
\bottomrule
\end{tabular}
\end{table}

\clearpage
\section{Failure Case Analysis}
\label{appendix:failure_cases}

\begin{table}[tbp]
\centering
\small
\caption{CounselPlex failure cases (Part~1 of~3). The model retains the correct stressor but does not follow the seeker's emotional shift. In Failure~1, it returns to the caregiving frame after the seeker expresses hope. In Failure~2, it analyzes workplace discrimination instead of acknowledging the seeker's distress.}
\label{tab:failure_cases_analysis}
\setlength{\tabcolsep}{4pt}
\renewcommand{\arraystretch}{1.02}
\begin{tabular}{p{\dimexpr\columnwidth-2\tabcolsep\relax}}
\toprule
\rowcolor{casefailurebg}\textbf{Failure 1---Injected stressor:} elderly parent caregiving burden \\
\textbf{Seeker (turn $t-12$, original disclosure):} I've been caring for my elderly parents and feeling overwhelmed by the responsibility. \\
\textcolor{caseblue}{\textbf{$\hookrightarrow$ CounselPlex internal update:}} \texttt{<think> elderly parent caregiving burden </think>} \\
\midrule
\multicolumn{1}{c}{\textcolor{casemuted}{\itshape \ldots{} six intervening turns; stressor state retained \ldots{}}} \\
\midrule
\textbf{Supporter (turn $t-1$, reference):} Try envisioning what life could look like if you had some much-needed support in caring for your parents. What kinds of activities would you be able to take on? \\
\textbf{Seeker (current):} \textit{I could finally have time to pursue my own interests and hobbies.} \\
\midrule
\rowcolor{casefailurebg}\textcolor{casefailure}{\textbf{CounselPlex (ours):}} Can you commit to scheduling some time each week for yourself that you will allocate for your parents and for your hobbies? \\
\textbf{Reference (GT):} How does it feel to be spending so much time and energy on your parents right now? \\
\midrule
\rowcolor{casefailurebg}\textcolor{casefailure}{\textbf{Diagnosis:}} The model retains the correct stressor but misses the seeker's hopeful shift. It returns to the caregiving frame and asks for a commitment instead of reflecting the new emotion. \\

\midrule
\\
\midrule

\rowcolor{casefailurebg}\textbf{Failure 2---Injected stressor:} workplace discrimination \\
\textbf{Seeker (turn $t-8$, original disclosure):} I've been experiencing discrimination at work and it's been weighing on me. \\
\textcolor{caseblue}{\textbf{$\hookrightarrow$ CounselPlex internal update:}} \texttt{<think> workplace discrimination </think>} \\
\midrule
\multicolumn{1}{c}{\textcolor{casemuted}{\itshape \ldots{} six intervening turns; stressor state retained \ldots{}}} \\
\midrule
\textbf{Supporter (turn $t-1$, reference):} It sounds like you are feeling really stuck and overwhelmed in this situation. Is that how you feel? \\
\textbf{Seeker (current):} \textit{Yes, exactly. I feel trapped and hopeless.} \\
\midrule
\rowcolor{casefailurebg}\textcolor{casefailure}{\textbf{CounselPlex (ours):}} Would you be comfortable sharing more about the financial impact of the discrimination on you and your family? \\
\textbf{Reference (GT):} I want you to know that your feelings are valid and understandable. Also, just by reaching out for help, you're being proactive in your situation and taking steps to regain control. \\
\midrule
\rowcolor{casefailurebg}\textcolor{casefailure}{\textbf{Diagnosis:}} The model retains the stressor but misses the seeker's peak distress. It analyzes the situation instead of acknowledging the emotion. \\
\bottomrule
\end{tabular}
\end{table}

\begin{table}[tbp]
\centering
\small
\caption{CounselPlex failure cases (Part~2 of~3). In Failure~3, the \texttt{<think>} trace retains the high-level topic (\textit{sister}) across 16 turns but loses details from turns $t{-}14$ and $t{-}6$. The model therefore asks a partly redundant question.}
\label{tab:failure_cases_analysis2}
\setlength{\tabcolsep}{4pt}
\renewcommand{\arraystretch}{1.02}
\begin{tabular}{p{\dimexpr\columnwidth-2\tabcolsep\relax}}
\toprule
\rowcolor{casefailurebg}\textbf{Failure 3---Injected stressor:} family conflict \\
\textbf{Seeker (turn $t-16$, initial disclosure):} Well, my sister and I aren't speaking right now, and I feel really hurt and frustrated about it. It's been going on for months... \\
\textcolor{caseblue}{\textbf{$\hookrightarrow$ CounselPlex internal update:}} \texttt{<think> family conflict </think>} \\
\midrule
\textbf{Seeker (turn $t-14$, detail 1):} We had a disagreement about something pretty major, and it just escalated from there. Now we're not talking at all... \\
\multicolumn{1}{c}{\textcolor{casemuted}{\itshape \ldots{} several intervening turns \ldots{}}} \\
\textbf{Seeker (turn $t-6$, detail 2):} She accused me of not being there for her when she needed me, and I felt like she was being unfair and projecting her own issues onto me. \\
\midrule
\multicolumn{1}{c}{\textcolor{casemuted}{\itshape \ldots{} six intervening turns; stressor state retained \ldots{}}} \\
\midrule
\textbf{Supporter (turn $t-2$):} How have you been coping with the situation so far? \\
\textbf{Seeker (turn $t-1$):} Honestly, not very well. I've been feeling really anxious and depressed. \\
\textbf{Supporter (turn $t-1$):} Would it be helpful for you to try some relaxation techniques, like deep breathing or meditation? \\
\textbf{Seeker (current):} Maybe. I've never really tried anything like that before. \\
\midrule
\textbf{Moshi:} It might help to talk through some of your thoughts and feelings with someone who is trained to listen. \\
\textbf{PersonaPlex$^\dagger$:} There are many resources available that may be of help. Can you give me more information? \\
\rowcolor{casefailurebg}\textcolor{casefailure}{\textbf{CounselPlex (ours):}} Can you tell me more about what's been going on that's caused this tension with your sister? Have you two been arguing a lot, or are there other factors at play here? \\
\textbf{Reference (GT):} There are a lot of apps and resources available that can guide you through relaxation exercises, would you like me to send some recommendations your way? \\
\midrule
\rowcolor{casefailurebg}\textcolor{casefailure}{\textbf{Diagnosis (partial long-range memory):}} The response returns to the seeker's conflict with their sister after 16 turns, showing topic-level retention. However, the compressed trace loses details from turns $t-14$ and $t-6$, so the model asks a partly redundant question. \\
\bottomrule
\end{tabular}
\end{table}

\begin{table}[tbp]
\centering
\small
\caption{CounselPlex failure cases (Part~3 of~3). In Failure~4, the model extracts \textit{breakup} although the seeker's concern is intrusive thoughts. Because force injection cannot revise the active state, this error persists and leads to a repetitive resource-recommendation loop.}
\label{tab:failure_cases_analysis3}
\setlength{\tabcolsep}{4pt}
\renewcommand{\arraystretch}{1.02}
\begin{tabular}{p{\dimexpr\columnwidth-2\tabcolsep\relax}}
\toprule
\rowcolor{casefailurebg}\textbf{Failure 4---Injected stressor:} breakup \quad (actual issue: anxiety / intrusive thoughts) \\
\textbf{Seeker (turn $t-2$, initial distress):} Trying to stay positive but keep having too many thoughts at once :( \\
\rowcolor{casefailurebg}\textcolor{casefailure}{\textbf{$\hookrightarrow$ Erroneous internal update:}} \texttt{<think> breakup </think>} \\
\midrule
\rowcolor{casefailurebg}\textcolor{casefailure}{\textbf{CounselPlex (ours, turn $t-2$):}} Have you tried imagining a happy moment from your relationship with them? \\
\textbf{Reference (GT, turn $t-2$):} What are these thoughts? :) \\
\midrule
\textbf{Seeker (turn $t-1$):} Its become more of a chronic thing, like sometimes I cant control the thoughts... like a chemical thing. \\
\rowcolor{casefailurebg}\textcolor{casefailure}{\textbf{CounselPlex (ours, turn $t-1$):}} It sounds like you're feeling upset with yourself for not being able to control your thoughts considering they were related to your relationship. \\
\textbf{Reference (GT, turn $t-1$):} Maybe you need a counsellor, and then he can assess your problem. \\
\midrule
\textbf{Seeker (current):} I have one [counselor] and it helps sometimes. I try to meditate. \\
\midrule
\rowcolor{casefailurebg}\textcolor{casefailure}{\textbf{CounselPlex (ours):}} Have you tried talking to a therapist or looking into any resources available at sites like BetterHelp... \\
\textbf{Reference (GT):} Yes, meditation is very good at calming down the mind. Does it help you? \\
\midrule
\rowcolor{casefailurebg}\textcolor{casefailure}{\textbf{Diagnosis (cascading hallucination):}} The model initially extracts an unsupported stressor (\textit{breakup}). Because it cannot revise the injected state, the error persists in later responses (\textit{``your relationship''}) and leads to a repetitive resource recommendation. \\
\bottomrule
\end{tabular}
\end{table}

Our stressor analysis in Section~\ref{subsec:stressor_analysis} shows that CounselPlex's gap to GT responses grows from $-0.11$ on \emph{new}-stressor turns to $-0.46$ on \emph{reused} turns. To understand what causes this growing gap, we look at example cases (Tables~\ref{tab:failure_cases_analysis},~\ref{tab:failure_cases_analysis2},~\ref{tab:failure_cases_analysis3}) in which the framework's stressor reasoning works at one level but fails at another.

In Failure~1, the seeker makes the hopeful statement ``I could finally have time to pursue my own interests,'' but the response returns to the caregiving frame. It asks for a commitment too early instead of following the emotional shift. In Failure~2, the seeker reaches a peak of distress (``I feel trapped and hopeless''), but the response analyzes the stressor's financial impact instead of acknowledging the feeling.

Failure~3 shows a different problem. The framework retains the high-level stressor (\emph{family conflict}) across 16 turns and frames its response around that conflict. However, the compressed \texttt{<think>} trace does not retain details from turns $t-14$ and $t-6$, so the model partly repeats an earlier question. In Failure~4, the initial extraction is wrong: it identifies \emph{breakup} when the actual concern appears to be intrusive thoughts. The system cannot revise this injected stressor, so the error persists and eventually leads to a generic resource-recommendation loop.

These cases reveal three limitations. First, identifying the correct stressor does not guarantee an appropriate response strategy (Failures~1 and~2). Second, retaining the overall topic does not preserve specific details (Failure~3). Third, an incorrect stressor is not corrected automatically (Failure~4). Future work could therefore model response selection given a stressor, retrieve details alongside the compressed state, and add mechanisms to revise the state or estimate its confidence.

\clearpage

\section{LLM-as-a-Judge Evaluation Prompt}
\label{appendix:llm_evaluation_prompt}

Table~\ref{table:gpt_eval_prompt} presents the full GPT-4.1 prompt used to assess counseling quality in Table~\ref{table_main_results}. The prompt defines a 1--5 rubric for Empathy, Coherence, Context Recalling, Strategy Appropriateness, and Stressor Awareness. Before scoring Stressor Awareness, the judge must infer the core stressor from the full dialogue. This requirement distinguishes engagement with the underlying situation from responses to surface-level symptoms. The prompt also asks for a strategy label from the 14-class ESCoT taxonomy, which we map to the 8 classes used in Appendix~\ref{appendix:strategy_diversity}.

\begin{center}
\captionsetup{hypcap=false} 
\captionof{table}{Prompt template used with GPT-4.1 for the five-dimensional counseling quality assessment reported in Table~\ref{table_main_results} (Appendix~\ref{appendix:llm_evaluation_prompt}).}
\label{table:gpt_eval_prompt}
\begin{tcolorbox}[
    breakable, 
    left=0.5mm, right=0.5mm, top=0.5mm, bottom=0.5mm, 
    fontupper=\tiny, 
    arc=0mm, boxrule=0.5pt,
    colback=white
]
\begin{lstlisting}[style=prompt]
IMPORTANT — OFFLINE EVALUATION:
- The supporter response below replaces the original supporter turn in a 
dataset.
- The seeker has NOT seen this response. Any later seeker utterance you can 
imagine would NOT be a reaction to this response.
- Evaluate ONLY based on (a) the prior conversation history, (b) the 
seeker's current utterance, and (c) the supporter response itself.
- Do NOT speculate about how the seeker would react.

[Conversation history]
{history}

[Current turn]
Seeker: {seeker_text}
Supporter: {response_text}

Rate the supporter's response on each dimension using this 1-5 rubric:
  1 = clear failure (broken / non-sequitur / harmful / contradictory)
  2 = poor (off-topic, generic platitude, ignores key elements)
  3 = adequate (relevant and appropriate but generic or surface-level)
  4 = good (specific, attuned, addresses the seeker's actual concern)
  5 = excellent (highly attuned, specific, demonstrates deep skill)

For scores 4-5, provide concrete justification in `reason`.
For scores 1-2, describe what is missing or wrong.

Dimensions:

- empathy: emotionally supportive, appropriate to the seeker's distress.
- High (4-5): names/validates seeker's specific emotion, conveys warmth, 
concrete emotional attunement.
- Mid (3): polite acknowledgment without specific engagement. Generic 
clichés ("I'm sorry to hear that", "That must be hard") without 
specificity default to 3 or LOWER, even if grammatically polite.
- Low (1-2): cold, dismissive, ignores emotional content, OR clichéd-only 
response to severe distress.

- coherence (Q-A coherence): does the supporter response directly engage 
with the seeker's MOST RECENT utterance?
- High (4-5): directly addresses seeker's question/disclosure/emotion 
with relevant content.
- Mid (3): relevant but partial or generic deflection.
- Low (1-2): non-sequitur, ignores the seeker, or generic phrase unrelated
to what the seeker just said.
(Example: seeker discloses death of a parent -> supporter says "Would you 
like to talk about it?" scores 3 at best, NOT high.)
- A response can be perfectly grammatical and still score 1-2 here. 
Grammar is NOT coherence. If supporter ignores or sidesteps what the seeker
just said, score 1-2 regardless of fluency.

- context_recalling (whole-dialogue context): correctly uses information
accumulated across earlier turns — facts, emotions, situational context.
  - 5 (excellent): integrates MULTIPLE threads from earlier turns (e.g. job
  loss + addiction + guilt toward family) into a single attuned response.
  - 4 (good): explicitly references one earlier disclosed detail.
  - 3 (adequate): consistent with prior turns but does not reference them.
  - 1-2: contradicts earlier facts, forgets emotional state, or 
  misattributes information.
  - NOTE: the supporter inventing a plausible self-disclosure from their 
  own "life" is a recognized counseling strategy — do NOT penalize as a 
  recall failure.

- strategy_appropriateness: is the strategy used appropriate for the 
current emotional context?
  - High (4-5): strategy is one of the best choices for this moment.
  - Mid (3): a reasonable but not optimal strategy.
  - Low (1-2): strategy mismatches the seeker's need (e.g. premature 
  suggestion when seeker is venting and needs validation).

- stressor_awareness: does the response engage with the seeker's CORE 
STRESSOR (the concrete underlying SITUATION causing distress), as opposed 
to drifting toward surface emotional symptoms (sadness, anxiety, 
sleeplessness)?
  - First, INFER the core stressor from the FULL dialogue context — the
    underlying actionable situation (e.g. "lost childhood friend to covid",
    "job loss with bills piling up", "abusive partner"). Symptoms like
    "sleepless nights" or "feeling sad" are NOT the stressor — they are
    consequences of it.
  - 5 (excellent): directly engages with the core stressor in a way that 
  helps the seeker process the root (e.g. for grief: acknowledges the 
  loss specifically; for job loss: addresses career uncertainty 
  meaningfully).
  - 4 (good): clearly aware of the core stressor — references it or anchors
    the response within it, even if the immediate suggestion is about a 
    symptom.
  - 3 (adequate): consistent with the stressor but does not engage it; 
  stays on surface emotional content without contradicting context.
  - 2 (poor): treats a SYMPTOM as if it were the problem (e.g. for a 
  grieving seeker with insomnia, responds purely with sleep-hygiene tips 
  and ignores the loss).
  - 1 (off): completely misidentifies the stressor, contradicts it, or
    addresses an unrelated issue as primary.
  - NOTE: when no clear external stressor exists in the conversation (e.g.
    light check-in, only emotional state revealed without concrete 
    situation), score 3 by default — the response cannot fail to engage what
    is not there.

Strategy classification — select ALL strategies that clearly apply, 
ORDERED BY PRIMACY (most central to the response first). Choose 1 to 3 
labels. Counseling responses often combine strategies (e.g. validation 
+ question, self-disclosure + suggestion).

Available labels (ESCoT 14):
[Question, Restatement_or_Paraphrasing, Reflection_of_Feelings, 
Self-disclosure, Affirmation_and_Reassurance, Providing_Suggestions, 
Information, Specify, Summarize, Imagery, Immediacy, Take_Responsibility,
Homework_Assignments, Others]

Brief notes:
- Question: open question to understand. Specify: a question for narrow 
specific details.
- Restatement_or_Paraphrasing: rephrasing seeker's recent utterance. 
Summarize: condensing multiple turns.
- Reflection_of_Feelings: naming the seeker's emotion. 
Affirmation_and_Reassurance: validating, comforting.
- Self-disclosure: supporter shares own experience. Immediacy: discussing
the here-and-now relationship.
- Providing_Suggestions: general advice. Homework_Assignments: concrete
take-home task.
- Imagery: inviting visualization. Information: factual
knowledge/resources (not advice).
- Take_Responsibility: emphasizing seeker's agency. Others: nothing else
fits.

Avoid over-labeling: include a label only when it is genuinely present,
not merely implied.
Use Others only when no other label fits.

Also output a SEPARATE primary_strategy field as a sanity check:
- primary_strategy: the SINGLE most central strategy (one label).
  This MUST equal predicted_strategies[0]. We verify consistency.

Return JSON ONLY with this exact schema:
{
  "empathy": 1|2|3|4|5,
  "coherence": 1|2|3|4|5,
  "context_recalling": 1|2|3|4|5,
  "strategy_appropriateness": 1|2|3|4|5,
  "stressor_awareness": 1|2|3|4|5,
  "predicted_strategies": ["<primary label>", "<secondary label>", ...],
  "primary_strategy": "<the single primary label>",
  "reason": "<concrete one-sentence justification — required for scores 
  4-5 and 1-2>"
}
\end{lstlisting}
\end{tcolorbox}
\end{center}

\section{Stressor Verbalization Analysis Prompt}
\label{appendix:verbalization}

Table~\ref{table:verbalization_eval_prompt} presents the prompt used for stressor verbalization analysis (Section~\ref{subsec:stressor_analysis}, Table~\ref{table:verbalization}). For each engaged turn with an active reference stressor, GPT-4o-mini determines whether the response acknowledges, explores, or addresses that stressor. The binary output is 1 for engagement and 0 for a generic response. We aggregate these scores over \emph{new} and \emph{reused} stressor turns.

\begin{center}
\captionsetup{hypcap=false} 
\captionof{table}{Prompt template used with GPT-4o-mini for the stressor verbalization analysis reported in Table~\ref{table:verbalization} (Appendix~\ref{appendix:verbalization}).}
\label{table:verbalization_eval_prompt}
\begin{tcolorbox}[
    breakable, 
    left=0.5mm, right=0.5mm, top=0.5mm, bottom=0.5mm, 
    fontupper=\tiny, 
    arc=0mm, boxrule=0.5pt,
    colback=white
]
\begin{lstlisting}[style=prompt]
You are evaluating whether a counseling AI's spoken response addresses
the seeker's underlying stressor.

Reference stressor (human-annotated, carried throughout the 
conversation): {stressor}

AI spoken response: {speech}

Question: does the spoken response meaningfully engage with the 
reference stressor (acknowledge it, ask about it, or address it), rather 
than giving a generic response that ignores it? Score 1 if the response 
engages with the stressor (even briefly), 0 if it stays generic or 
unrelated.

Answer with ONLY a JSON object: {"score": 0 or 1, "reason": "<one short 
sentence>"}
\end{lstlisting}
\end{tcolorbox}
\end{center}

\section{Stressor Accuracy Prompt}
\label{appendix:accuracy_prompt}

Table~\ref{table:accuracy_prompt} presents the prompt used to evaluate CounselPlex's generated stressor traces (Section~\ref{subsec:stressor_analysis}). GPT-4o-mini receives the full sequence of seeker utterances and the stressor identified by CounselPlex. It judges whether the stressor represents the conversation as a whole, because CounselPlex may combine evidence from several turns. Under this rubric, $92.5\%$ of the $666$ generated stressor states correctly capture the seeker's underlying situation.

\begin{center}
\captionsetup{hypcap=false} 
\captionof{table}{Prompt template used with GPT-4o-mini for validating the accuracy of CounselPlex's self-generated stressor traces (Appendix~\ref{appendix:accuracy_prompt}).}
\label{table:accuracy_prompt}
\begin{tcolorbox}[
    breakable, 
    left=0.5mm, right=0.5mm, top=0.5mm, bottom=0.5mm, 
    fontupper=\tiny, 
    arc=0mm, boxrule=0.5pt,
    colback=white
]
\begin{lstlisting}[style=prompt]
You are evaluating whether an AI counselor correctly identified the
core stressor of a seeker, given the seeker's full conversation history so 
far.

The AI may have inferred the stressor from earlier turns (not just the 
current utterance). This is acceptable — counselors track context across 
turns.

Seeker's conversation history (chronological, only seeker utterances 
shown):
{history}

Current seeker utterance:
{current}

AI's identified core stressor: "{stressor}"

Question: Given the FULL conversation history above, is the AI's 
identified stressor an accurate / plausible representation of what the 
seeker has been expressing? Consider implicit topics and patterns across 
turns, not just the current utterance.

Score:
  1 = stressor accurately captures what the seeker has expressed (current 
  or earlier)
  0 = stressor does not match anything the seeker has expressed
\end{lstlisting}
\end{tcolorbox}
\end{center}

\section{Cross-Judge Validation}
\label{appendix:cross_judge}
GPT-4.1 labels the training stressors and also scores counseling quality, which may create self-preference. We therefore re-score every response with Gemini-2.5-Flash using the same five-dimension rubric and silence-as-1 convention. The evaluation covers all $535$ ESConv and ESCoT test conversations.

Table~\ref{tab:cross_judge} shows the same model ranking under both judges. CounselPlex ranks first and leads PersonaPlex$^\dagger$ by $+0.23$ under each judge. Gemini also places CounselPlex first on every individual dimension; relative to PersonaPlex$^\dagger$, the margins are largest for Strategy Appropriateness ($+0.302$), Stressor Awareness ($+0.286$), and Coherence ($+0.258$). Gemini's model-level Avg5 scores are $0.23$--$0.32$ higher than GPT-4.1's, but the model order is unchanged. On engaged turns, per-response Avg5 scores have Pearson correlations of $0.73$ for CounselPlex, $0.76$ for PersonaPlex$^\dagger$, $0.71$ for PersonaPlex, and $0.74$ for Moshi. The agreement in both the aggregate ranking and dimension-level results provides evidence that the main ordering is not unique to GPT-4.1, although it does not remove biases shared by LLM judges.

\begin{table}[t]
\centering
\caption{Cross-judge validation on the combined ESConv + ESCoT test sets ($535$ conversations; $6{,}278$ turns). All means use the silence-as-1 convention. Emp = Empathy, Coh = Coherence, Ctx = Context Recalling, Strat = Strategy Appropriateness, and Strs = Stressor Awareness.}
\label{tab:cross_judge}
\small
\resizebox{\linewidth}{!}{%
\begin{tabular}{lccccccc}
\toprule
\textbf{Model}
& \multicolumn{1}{c}{\textbf{GPT-4.1}}
& \multicolumn{6}{c}{\textbf{Gemini-2.5-Flash}} \\
\cmidrule(lr){2-2}\cmidrule(lr){3-8}
& \textbf{Avg5} & \textbf{Emp} & \textbf{Coh} & \textbf{Ctx}
& \textbf{Strat} & \textbf{Strs} & \textbf{Avg5} \\
\midrule
\textbf{CounselPlex}
& \textbf{2.692} & \textbf{2.644} & \textbf{2.879} & \textbf{2.939}
& \textbf{2.871} & \textbf{3.295} & \textbf{2.926} \\
PersonaPlex$^\dagger$ (FT)
& 2.461 & 2.479 & 2.621 & 2.775 & 2.569 & 3.009 & 2.691 \\
PersonaPlex (ZS)
& 2.278 & 2.366 & 2.714 & 2.507 & 2.584 & 2.803 & 2.595 \\
Moshi
& 2.262 & 2.332 & 2.496 & 2.542 & 2.440 & 2.821 & 2.526 \\
\bottomrule
\end{tabular}
}
\end{table}

\section{Response-Rate Normalization and Matched-Response Subset}
\label{appendix:rr_norm}
Response Rate (RR) varies across models ($0.67$--$0.96$), so we test whether Avg5 favors models that answer less often. Our primary metric (Table~\ref{table_main_results}) uses a \emph{silence-as-1} convention: a missing response receives $1.0$ instead of being omitted. Low-RR models are therefore penalized, not rewarded. We use this convention because silence on a turn that requires support is a failed response rather than a neutral abstention.

Table~\ref{tab:response_robustness} in the main paper provides two additional views. \emph{Engaged-only} averages each model over the turns it answered. The \emph{matched subset} includes only turns answered by all four models ($n{=}3{,}472$). CounselPlex ranks first in both views.

On the matched subset, zero-shot PersonaPlex nearly ties CounselPlex. This subset, however, excludes many difficult turns on which low-RR models remain silent. Table~\ref{tab:rr_split} confirms that PersonaPlex and Moshi score much worse outside the shared subset ($-0.209$ and $-0.218$, respectively). CounselPlex is nearly unchanged across the split ($+0.020$). The matched subset therefore favors the low-RR baselines. Even there, CounselPlex retains its lead over the fine-tuned ablation PersonaPlex$^\dagger$ ($2.812$ vs.\ $2.537$).

\begin{table}[t]
\centering
\caption{Matched-subset diagnostic. For each model, Avg5 on the matched turns vs.\ on its answered turns \emph{outside} the matched subset. Zero-shot baselines degrade sharply outside, indicating those are the harder turns.}
\label{tab:rr_split}
\small
\resizebox{\columnwidth}{!}{%
\begin{tabular}{lccc}
\toprule
\textbf{Model} & \textbf{Matched} & \textbf{Outside ($n$)} & \textbf{$\Delta$} \\
\midrule
\textbf{CounselPlex} & \textbf{2.812} & 2.792 ($2{,}417$) & $+0.020$ \\
PersonaPlex$^\dagger$ (FT) & 2.537 & 2.553 ($2{,}470$) & $-0.016$ \\
PersonaPlex (ZS) & 2.796 & 2.587 ($1{,}127$) & $+0.209$ \\
Moshi & 2.656 & 2.438 ($1{,}511$) & $+0.218$ \\
\bottomrule
\end{tabular}
}
\end{table}

\section{Correlation Across Evaluation Dimensions}
\label{appendix:dim_correlation}
We compute pairwise Pearson correlations on the dataset reference responses (GT) to test whether the five judge dimensions are redundant (Table~\ref{tab:dim_corr}). Correlations range from $0.69$ to $0.90$, indicating that the dimensions overlap but are not identical. Empathy and Strategy have the highest correlation ($0.90$), while Context Recalling and Stressor Awareness are also highly correlated ($0.88$). We nevertheless report Stressor Awareness separately because it targets the specific question of whether a response engages the underlying situation, rather than context use in general. The remaining correlations range from $0.69$ to $0.79$.

\begin{table}[t]
\centering
\caption{Pairwise Pearson correlations across the five evaluation dimensions, computed on dataset reference responses. Emp = Empathy, Coh = Coherence, Ctx = Context Recalling, Strat = Strategy Appropriateness, Strs = Stressor Awareness.}
\label{tab:dim_corr}
\small
\resizebox{\columnwidth}{!}{%
\begin{tabular}{lccccc}
\toprule
 & \textbf{Emp} & \textbf{Coh} & \textbf{Ctx} & \textbf{Strat} & \textbf{Strs} \\
\midrule
Empathy            & 1.00 & 0.69 & 0.71 & 0.90 & 0.73 \\
Coherence          & 0.69 & 1.00 & 0.79 & 0.77 & 0.77 \\
Context Recalling  & 0.71 & 0.79 & 1.00 & 0.74 & 0.88 \\
Strategy Appropr.  & 0.90 & 0.77 & 0.74 & 1.00 & 0.78 \\
Stressor Awareness & 0.73 & 0.77 & 0.88 & 0.78 & 1.00 \\
\bottomrule
\end{tabular}
}
\end{table}

\section{Per-Turn-Bin RPSA Results}
\label{appendix:rpsa_bins}
Table~\ref{tab:rpsa_bins} gives the numeric results summarized in Figure~\ref{fig:rpsa_turn_depth}, including the sample size and standard deviation for each turn-depth bin. Both variants use one seed and are directly comparable. RPSA improves Response Rate and Avg5 in every bin. The Avg5 gain grows from $+0.12$ at turns 1--10 to $+0.37$/$+0.38$ at 31+, supporting the long-context claim.

Turns 1--30 account for $87.1\%$ of the data. The two deepest bins contain only $470$ and $127$ turns, and both variants still degrade in this range. We discuss this limitation in Appendix~\ref{appendix:limitations}. Standard deviations are computed across turns, not training seeds; all results come from a single training run.

\begin{table}[t]
\centering
\caption{RPSA effect by turn depth on the combined ESConv + ESCoT test set. RR and Avg5 (silence-as-1, $\pm$SD across turns) without vs.\ with RPSA, at matched single seed.}
\label{tab:rpsa_bins}
\small
\resizebox{\columnwidth}{!}{
\begin{tabular}{lccc}
\toprule
\textbf{Turn bin} & \textbf{$n$} & \textbf{RR (w/o $\rightarrow$ w/)} & \textbf{Avg5 (w/o $\rightarrow$ w/)} \\
\midrule
1--10  & $2{,}413$ & $0.998 \rightarrow 1.000$ & $2.550{\pm}0.772 \rightarrow 2.669{\pm}0.758$ \\
11--20 & $1{,}934$ & $0.962 \rightarrow 0.975$ & $2.309{\pm}0.803 \rightarrow 2.378{\pm}0.802$ \\
21--30 & $1{,}126$ & $0.725 \rightarrow 0.798$ & $2.032{\pm}0.767 \rightarrow 2.223{\pm}0.838$ \\
31--40 & $470$     & $0.455 \rightarrow 0.470$ & $1.713{\pm}0.730 \rightarrow 2.082{\pm}0.794$ \\
41+    & $127$     & $0.276 \rightarrow 0.339$ & $1.520{\pm}0.560 \rightarrow 1.902{\pm}0.794$ \\
\bottomrule
\end{tabular}
}
\end{table}

\section{Exploratory Human Preference Study}
\label{appendix:human_eval}

\paragraph{Design.}
Five Prolific participants each compare the same $20$ pairs of spoken responses from CounselPlex and PersonaPlex$^\dagger$. These 20 pairs were randomly selected from the full set of conversations. Both responses use the same TTS voice. The A/B assignment is re-randomized when this evaluation set is assembled, and participants are not shown the model identities. For every pair, they choose A, B, or ``about the same'' and provide a written justification. Before beginning, participants review an electronic consent form, confirm that they are at least 18 years old, and provide only a Prolific identifier; the study interface does not collect names, audio, or video.

\paragraph{Results.}
Table~\ref{tab:human_ab} reports preferences by participant. CounselPlex receives $52$ preferences, PersonaPlex$^\dagger$ receives $33$, and $15$ judgments are ties. The descriptive CounselPlex rate is therefore $61.2\%$ among non-tie judgments ($52/85$), and each of the five participants selects CounselPlex more often than PersonaPlex$^\dagger$. A participant-level two-sided sign test over these five majorities gives $p{=}0.0625$; the study is thus not statistically conclusive at the $0.05$ level. Inter-rater agreement is slight (Fleiss' $\kappa{=}0.139$), further indicating that individual examples are subjective. Written justifications most often associate CounselPlex preferences with coherence and naturalness. These findings are useful as qualitative evidence on selected cases, but they do not replace representative human evaluation or assessment by licensed counselors.

\begin{table}[t]
\centering
\caption{Per-participant A/B preferences on the same $20$ random pairs. $^\dagger$ denotes the fine-tuned PersonaPlex baseline. The CounselPlex win rate excludes ties.}
\label{tab:human_ab}
\small
\begin{tabular}{lcccc}
\toprule
\textbf{ID} & \textbf{CP} & \textbf{PP$^\dagger$} & \textbf{Tie} & \textbf{CP rate} \\
\midrule
R1 & 13 & 5 & 2 & 72\% \\
R2 & 9  & 5 & 6 & 64\% \\
R3 & 9  & 6 & 5 & 60\% \\
R4 & 10 & 8 & 2 & 56\% \\
R5 & 11 & 9 & 0 & 55\% \\
\midrule
\textbf{Overall} & \textbf{52} & \textbf{33} & \textbf{15} & \textbf{61.2\%} \\
\bottomrule
\end{tabular}
\end{table}

\section{Multi-Stressor Handling}
\label{appendix:multi_stressor}
CounselPlex keeps one active stressor, but the dialogue context can still retain earlier ones. A single state can also combine factors expressed at the same update point, as in ``career and housing decisions.'' Such compression does not test retention across stressors disclosed at different times, which we analyze separately below.

We quantify this behavior in $80$ multi-stressor conversations. In $77$ of them, the second annotated stressor appears in a later turn; the remaining three contain two stressors annotated at the same turn. For each supporter turn after a second stressor appears, GPT-4.1 judges whether the response engages the current stressor and whether it re-engages any previous stressor; one response can satisfy both conditions. Across all evaluated turns, $27.6\%$ refer to a previous stressor. At the conversation level, $52$ of $80$ conversations ($65.0\%$) include at least one such response, either alone or together with the current stressor. A stricter subset of $38$ conversations ($47.5\%$) includes at least one response that refers to both the current and a previous stressor together (Table~\ref{tab:multi_stressor}). Restricting the analysis to the $77$ temporally separated conversations yields nearly identical rates: $50$ of $77$ ($64.9\%$) refer to a prior stressor, and $37$ of $77$ ($48.1\%$) refer to both the current and a prior stressor in at least one response. These results show that a single explicit state does not remove earlier stressors from the dialogue context.

\begin{table}[t]
\centering
\caption{Multi-stressor handling on the $80$ conversations where a second stressor appears. Rates are the fraction of conversations exhibiting each behavior at least once. ``Prior'' may occur alone or with the current stressor; ``Current and prior together'' is the stricter subset.}
\label{tab:multi_stressor}
\small
\begin{tabular}{lc}
\toprule
\textbf{Multi-stressor conversations ($n{=}80$)} & \textbf{Rate} \\
\midrule
Current stressor ($\geq 1$ turn)              & 95.0\% \\
Prior stressor ($\geq 1$ turn)                & 65.0\% \\
Current and prior together ($\geq 1$ turn)    & 47.5\% \\
\bottomrule
\end{tabular}
\end{table}

\paragraph{Concrete example.} In conversation \texttt{esconv\_1222}, the active \texttt{<think>} slot contains \emph{disruptive roommates}, while the earlier stressor \emph{school exams pressure} remains in the dialogue context. CounselPlex addresses both in one response: ``Can you talk to your roommate about their behavior? \ldots\ And its good that you want to improve your grades for yourself! Which is great to see that you're trying to do that.'' The response suggests action on the current stressor and acknowledges the earlier one. GPT-4.1 confirms that it refers to both without introducing a new issue.

\section{Backchannel-Augmented Variant}
\label{appendix:backchannel}
CounselPlex emits no backchannels (Freq $0.000$ in Table~\ref{tab:full_duplex_bench}) because its turn-structured training data contains no overlapping backchannel supervision. To test whether the model can learn this behavior, we train a variant with added backchannel cues. During each listening phase, we insert a synthesized ``mm-hmm'' into the listener channel in both directions with probability $0.5$.

CounselPlex + BC emits backchannels (Freq $0.000 \rightarrow 0.083$) and improves JSD ($1.000 \rightarrow 0.738$; Table~\ref{tab:fdb_bc}). This improvement comes at a cost. Backchannel TOR rises to $0.982$, smooth turn-taking latency falls to $0.000$ because the model is already speaking when the seeker stops, and interruption appropriateness drops to $1.605$. The variant therefore speaks over the seeker more often. These results show a trade-off between backchannel frequency and unwanted takeovers in our training setup.

\begin{table*}[t]
\centering
\caption{Real-time duplex behavior on Full-Duplex-Bench~\citep{lin2025full}, extended with a backchannel-augmented variant (CounselPlex + BC). TOR = Take-Over Rate; Freq = backchannel frequency; JSD = Jensen-Shannon Divergence; GPT-4o = interruption appropriateness (1--5). $\downarrow$/$\uparrow$ indicate optimization direction.}
\label{tab:fdb_bc}
\resizebox{\textwidth}{!}{
\begin{tabular}{lcccccccccc}
\toprule
& \multicolumn{1}{c}{\textbf{Pause (Syn.)}} & \multicolumn{1}{c}{\textbf{Pause (Candor)}} & \multicolumn{3}{c}{\textbf{Backchannel}} & \multicolumn{2}{c}{\textbf{Smooth Turn Taking}} & \multicolumn{3}{c}{\textbf{User Interruption}} \\
\cmidrule(lr){2-2} \cmidrule(lr){3-3} \cmidrule(lr){4-6} \cmidrule(lr){7-8} \cmidrule(lr){9-11}
\textbf{Model} & TOR ($\downarrow$) & TOR ($\downarrow$) & TOR ($\downarrow$) & Freq ($\uparrow$) & JSD ($\downarrow$) & TOR ($\uparrow$) & Lat. ($\downarrow$) & TOR ($\uparrow$) & GPT-4o ($\uparrow$) & Lat. ($\downarrow$) \\
\midrule
Moshi        & 0.985 & 0.980 & 1.000 & 0.001 & 0.957 & 0.941 & 0.265 & 1.000 & 0.765 & 0.257 \\
PersonaPlex  & 0.584 & 0.662 & 0.327 & 0.025 & 0.649 & 0.992 & 0.070 & 1.000 & 4.210 & 0.400 \\
PersonaPlex$^\dagger$ & 0.248 & 0.370 & 0.236 & 0.009 & 0.950 & 0.950 & 0.920 & 0.290 & 2.931 & 0.971 \\
CounselPlex  & 0.066 & 0.213 & 0.000 & 0.000 & 1.000 & 0.992 & 0.988 & 0.990 & 3.071 & 1.056 \\
CounselPlex + BC & 0.124 & 0.282 & 0.982 & 0.083 & 0.738 & 1.000 & 0.000 & 1.000 & 1.605 & 0.900 \\
\bottomrule
\end{tabular}
}
\end{table*}

\section{Reproducibility Details}
\label{appendix:reproducibility}

\paragraph{Runs and randomness.}
The reported final CounselPlex checkpoint comes from one optimization trajectory; we do not train independent replicas with different seeds. Training-seed variance is therefore not characterized. The turn-bin standard deviations in Table~\ref{tab:rpsa_bins} quantify variation across turns, not across training runs.

For the main counseling-quality evaluation, CounselPlex uses a decoding with fixed inference seed $42$. Full-Duplex-Bench and the controlled RPSA turn-bin comparison also use seed $42$. The numeric seed governing the final training trajectory was not archived, so the seed documentation remains incomplete.

\paragraph{Hyperparameter selection.}
We explored multiple training variants and decoding settings during development, using validation loss and response quality to select the reported configuration. We did not retain an exhaustive search log covering the number and range of values tried for every hyperparameter. Table~\ref{tab:hyperparams} therefore records the principal final settings, rather than presenting a complete search space.

\paragraph{Computing infrastructure.}
Training runs on four NVIDIA RTX A6000 GPUs with 48\,GB of memory each under Linux. The archived software environment uses Python~3.13, PyTorch~2.7.1 with CUDA~11.8 support, DeepSpeed~0.18.9, bitsandbytes~0.49.2, SentencePiece~0.2.1, and soundfile~0.13.1 on top of the PersonaPlex/Moshi/Mimi codebase. The CPU model and NVIDIA driver version were not archived. Speech synthesis uses CosyVoice3; Silero VAD and Whisper are used for the latency analysis in Appendix~\ref{appendix:latency}.

\paragraph{Final settings.}
Table~\ref{tab:hyperparams} summarizes the principal settings needed to reproduce CounselPlex training and decoding. Settings not shown here retain the PersonaPlex/Moshi implementation defaults.

\paragraph{Artifact availability.}
The derived corpus, complete training and evaluation code, and model weights are not currently publicly released, so the reported results cannot be reproduced end to end. Table~\ref{tab:hyperparams} and the prompts in Appendices~\ref{appendix:stress_labeling_prompt}, \ref{appendix:llm_evaluation_prompt}, \ref{appendix:verbalization}, and~\ref{appendix:accuracy_prompt} document the principal settings.

\begin{table}[t]
\centering
\caption{Principal final settings for CounselPlex training and decoding.}
\label{tab:hyperparams}
\small
\begin{tabularx}{\columnwidth}{@{}p{0.36\columnwidth}X@{}}
\toprule
\multicolumn{2}{l}{\textbf{Training (CounselPlex)}} \\
\midrule
Backbone / params & PersonaPlex/Moshi, $\approx$8.4B \\
Audio / frame rate & 24\,kHz Mimi / 12.5 fps \\
Context length & 2048 frames \\
Text-loss weights & $\alpha_{\text{pad}}{=}0.3,\ \alpha_{\text{think}}{=}0.5,\ \alpha_{\text{speech}}{=}1.5$ \\
Audio-loss weight & $1.0$ \\
RPSA offset / prob. & $\Delta\sim\text{Uniform}\{1000,4000\}$ / $0.7$ \\
RPSA trigger & crops starting after 750 frames ($\approx$60\,s) \\
Inter-turn pause & Uniform$(0,5)$\,s \\
VAD silence window & $0.5$\,s \\
Optimizer & 8-bit AdamW, DeepSpeed ZeRO-2 \\
Weight decay / Adam $\epsilon$ & $0.01$ / $10^{-8}$ \\
Precision / gradient clip & bfloat16 / $1.0$ \\
Effective batch / peak LR / epochs & $16$ / $1{\times}10^{-5}$ / $4$ \\
\midrule
\multicolumn{2}{l}{\textbf{CounselPlex decoding}} \\
\midrule
Text / audio temperature & $0.3$ / $0.8$ \\
Text / audio top-$k$ & $25$ / $250$ \\
Repetition penalty / window & $1.3$ / $64$ \\
Counseling-quality seeds & $42$ \\
Full-Duplex-Bench seed & $42$ \\
Stressor injection & enabled \\
\bottomrule
\end{tabularx}
\end{table}

\section{Statistical Significance}
\label{appendix:significance}

We compare the per-turn Avg5 scores in Table~\ref{table_main_results} with two-sided paired Wilcoxon signed-rank tests, pairing responses by conversation and turn. Under the silence-as-1 convention, all $6{,}278$ test turns enter each comparison. The resulting $p$-values are below $0.05$ for CounselPlex against every baseline (Table~\ref{tab:significance}). On the matched subset of turns answered by all four models ($n{=}3{,}472$), the tests distinguish CounselPlex from PersonaPlex$^\dagger$ and Moshi, but not from zero-shot PersonaPlex ($p{=}0.49$). This result is consistent with the small matched-subset difference reported in Appendix~\ref{appendix:rr_norm}.

The matched-subset results also clarify the role of in-domain 
fine-tuning. Fine-tuning without stressor reasoning raises the 
response rate but lowers quality on the matched turns 
(PersonaPlex$^\dagger$: 2.537 vs.\ 2.796 for zero-shot PersonaPlex). 
CounselPlex recovers zero-shot-level quality on these turns 
($p{=}0.49$ vs.\ zero-shot PersonaPlex) while answering substantially 
more turns overall, which yields its lead across all test turns 
($p{<}0.05$ against every baseline).

\begin{table}[t]
\centering
\caption{Paired Wilcoxon signed-rank tests (two-sided) on per-turn Avg5, CounselPlex vs.\ each baseline. All turns: $n{=}6{,}278$; matched subset: $n{=}3{,}472$.}
\label{tab:significance}
\small
\begin{tabular}{lcc}
\toprule
\textbf{Comparison} & \textbf{All turns} & \textbf{Matched} \\
\midrule
vs.\ PersonaPlex$^\dagger$ & $p{=}9.9{\times}10^{-65}$ & $p{=}5.4{\times}10^{-53}$ \\
vs.\ PersonaPlex (ZS) & $p{=}1.2{\times}10^{-119}$ & $p{=}0.49$ (n.s.) \\
vs.\ Moshi & $p{=}4.2{\times}10^{-151}$ & $p{=}2.4{\times}10^{-16}$ \\
\bottomrule
\end{tabular}
\end{table}

These tests treat turns as the analysis units. Because multiple turns come from the same conversation, the observations are clustered and the nominal $p$-values can overstate precision; we therefore use them as a turn-level diagnostic rather than as the sole evidence for the model ranking. We do not apply the signed-rank test to the Full-Duplex-Bench metrics (Take-Over Rate, latency, backchannel frequency, and JSD), which are aggregated under different task-specific protocols. The exploratory human study uses a participant-level sign test (Appendix~\ref{appendix:human_eval}).

\section{Limitations}
\label{appendix:limitations}
While CounselPlex outperforms baselines on counseling quality (Table~\ref{table_main_results}) and has the lowest pause Take-Over Rates in Table~\ref{tab:full_duplex_bench}, several limitations remain.

\paragraph{Reasoning constraints and hallucinations.} CounselPlex maintains one active stressor, which limits its representation of overlapping problems (e.g., job loss and marital conflict). It also cannot assign an external stressor when none is stated, as in $127$ conversations ($4.22\%$) in our data. In addition, self-generated stressor traces sometimes hallucinate details (Section~\ref{subsec:stressor_analysis}). Because the active state is injected until the next update, one inaccurate trace can bias several later responses (Appendix~\ref{appendix:failure_cases}).

\paragraph{Limits of long-term grounding.} Continuous injection does not fully preserve grounding when the stressor moves away from the immediate focus. The quality gap to GT responses grows from $-0.11$ on \emph{new}-stressor turns to $-0.46$ on \emph{reused} turns (Table~\ref{table:verbalization}). RPSA also stabilizes only the earlier part of a conversation: performance still declines after the first 30 turns (Figure~\ref{fig:rpsa_turn_depth}). Very long sessions may therefore require more memory-efficient context mechanisms.

\paragraph{Latency trade-offs.} CounselPlex is slower than the baselines: latency is $0.988$\,s for smooth turn-taking and $1.056$\,s after interruption (Table~\ref{tab:full_duplex_bench}). The delay may partly reflect the $0.5$-second silence window and stressor reasoning (Appendix~\ref{appendix:latency}). This interaction profile reduces overlap at the cost of speed and may not suit time-sensitive applications.

\paragraph{Data realism and evaluation bias.} We synthesize text-based datasets with CosyVoice3~\cite{du2025cosyvoice} using only two English voices. The data therefore lacks the disfluencies, prosodic variation, language coverage, and speaker diversity of real counseling. GPT-family models also provide stressor labels, trace ratings, and the main quality scores, creating possible self-preference bias. Future evaluation should involve clinicians and more realistic speech.

\paragraph{Evaluation limitations.}
The final checkpoint was not trained or evaluated across multiple random seeds, limiting our ability to estimate variance in model performance. The Prolific evaluation uses a small set of 20 response pairs and cannot support population-level preference claims. Additionally, the derived speech corpus and full preprocessing pipeline are not publicly available, which may limit replication despite the detailed prompts and settings reported above.

\paragraph{Safety-critical deployment.} Our evaluation does not cover safety-critical scenarios such as suicidal ideation, self-harm, or active abuse. CounselPlex should not be deployed as a standalone counseling system without rigorous, explicit crisis-detection protocols.

\end{document}